\documentclass{article}
\PassOptionsToPackage{numbers, compress}{natbib}
\usepackage[main, final]{neurips_2026}

\usepackage[utf8]{inputenc}
\usepackage[T1]{fontenc}
\usepackage{microtype}

\usepackage{amsmath}
\usepackage{amssymb}
\usepackage{amsfonts}
\usepackage{nicefrac}

\usepackage{booktabs}
\usepackage{array}
\usepackage{multirow}
\usepackage{etoc}

\usepackage{graphicx}
\usepackage{float}
\usepackage{wrapfig}
\usepackage{caption}

\usepackage{algorithm}
\usepackage{algpseudocode}

\usepackage[inline]{enumitem}
\usepackage{soul}
\usepackage{xcolor}

\usepackage{etoc}
\usepackage{url}
\usepackage[colorlinks=true,
            linkcolor=black,
            citecolor=blue!70!black,
            urlcolor=blue!70!black]{hyperref}

\etocsetstyle{section}
  {\par\vspace{0.45em}\noindent}
  {\bfseries}
  {\makebox[1.8em][l]{\etocnumber}\etocname\nobreak\leaders\hbox{\normalfont\bfseries.\hspace{0.4em}}\hfill\nobreak\etocpage\par}
  {}

\etocsetstyle{subsection}
  {\noindent\leftskip=1.8em}
  {\normalfont}
  {\makebox[2.2em][l]{\etocnumber}\etocname\nobreak\leaders\hbox{\normalfont.\hspace{0.4em}}\hfill\nobreak\etocpage\par}
  {}

\newcommand{\wildcard}[4][4.5em]{\makebox[#1][s]{#2\hfill#3\hfill#4}}
  
\title{\textbf{LEMON-ZEST: Evolution-Informed Tokenization\\for Efficient Protein Language Modeling}}

\author{%
  \begin{tabular*}{\textwidth}{@{\extracolsep{\fill}}ccc@{}}
    \textbf{Biswajit Banerjee}$^{1}$ & \textbf{Claudia Alvarez-Carre\~{n}o}$^{2}$ & \textbf{Anton S. Petrov}$^{1}$ \\[0.3ex]
    \texttt{\small sumon@gatech.edu} & \texttt{\small c.carreno@ucl.ac.uk} & \texttt{\small ap211@gatech.edu}
  \end{tabular*} \\[2.8ex]
  $^{1}$Georgia Institute of Technology \qquad\qquad $^{2}$University College London
}

\begin{document}

\maketitle

\begin{abstract}
Protein Language Models (PLMs) have made remarkable progress following scaling laws established in natural language processing across sequence- and structure-based tasks, yet the potential of tokenization remains underexploited. Unlike human language, proteins preserve structure despite extensive sequence variation,  a property standard tokenization strategies fundamentally fail to capture. We introduce ZEST (Zoned Encoding of Sequence Traits), an evolution-informed vocabulary derived from conserved regions of multiple sequence alignments. ZEST allows embedding domain-level biological priors directly at the tokenization stage rather than learning them implicitly through scale. ZEST natively compresses sequences to an average token length of 3.3 residues, enabling our model to process $\geq3380$ residues within a standard 1024-token context window. Building on this, we present LEMON (Layered Extraction of Molecular Ordering from Nature), a compact 200M-parameter sequence-based model for detection of remote homology between protein sequences trained on a single H100 GPU for one week. Despite its modest size, LEMON outperforms models of 600M to 3B parameters and is on-par with TM-Vec. Our results demonstrate that evolution-informed tokenization can substitute for massive parameter scaling, opening a new direction for efficient, biologically-grounded protein representation learning. All code and model weights are publicly available (at \href{https://github.com/LDWLab/ZEST}{github.com/LDWLab/ZEST}) under the MIT license.
% While structure prediction models solve the surface level problem mapping a given sequence to its most probably 3D-structure, we ask a much harder question the origin of the sequence/structure itself. Consequently, 
% Zoned Encoding of Sequence Traits (ZEST),
%  Layered Extraction of Molecular Ordering from Nature (LEMON), 

% todo modify the abstract to be on par
\end{abstract}

\vspace{-1.5em}
\begin{figure}[H]
    \centering
    \includegraphics[width=\textwidth]{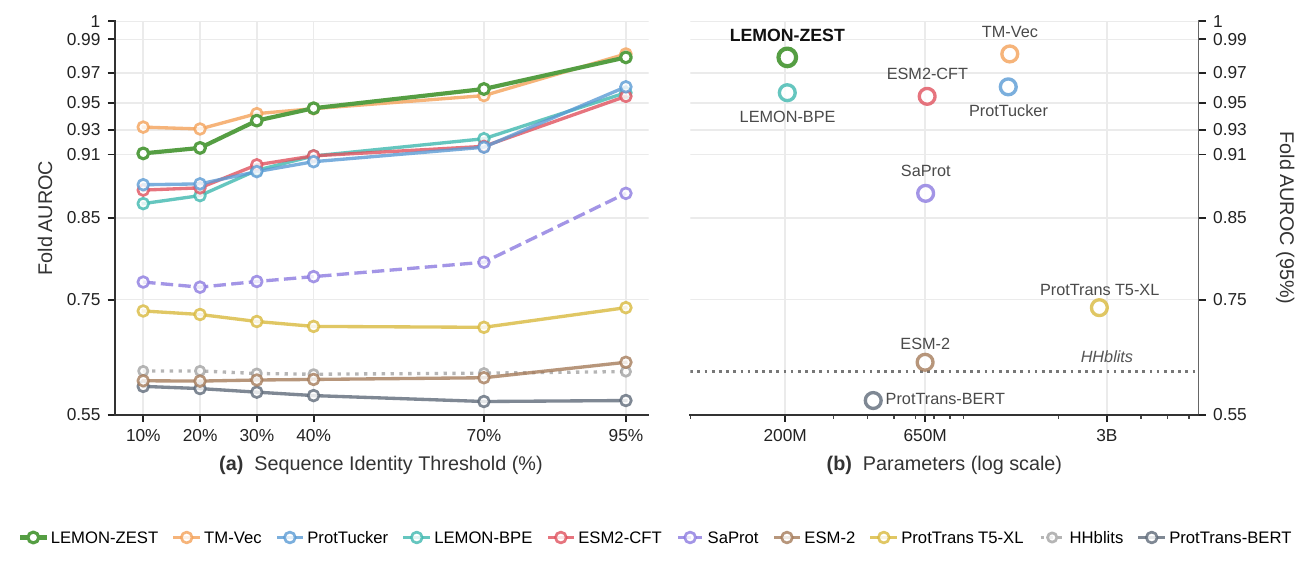}
    % \caption{(a) AUROC computed on SCOP dataset plotted against Sequence Identity Threshold; each point corresponds to an evaluation where the retrieval pool is filtered to proteins sharing at most t\% sequence identity with the query (x-axis). Dashed lines indicate models requiring 3Di structural tokens as input (SaProt, ProstT5) and a dotted line represents HHblits. (b) Mean fold-level AUROC averaged across all six thresholds in comparison to model size. HHblits is a non-parametric, HMM profile-based  comparison.}
    \caption{\small \textbf{SCOPe fold-level retrieval.} \textbf{(a)} Fold AUROC restricted to proteins sharing at most t\% sequence identity with the query. \textbf{(b)} Fold AUROC$@95\%$ versus parameter count (log scale). TM-Vec is trained with structural (TM-score) supervision. LEMON-BPE is the matched control with BPE tokenization,  ESM2-CFT is ESM-2 650M contrastively finetuned end to end with LEMON's representation head  and data. SaProt (dashed) utilizes structure-derived 3Di tokens. HHblits (dotted) relies on Hidden Markov Model-profile based search. }
    \label{fig:threholds}
\end{figure}

\section{Introduction}
The success of modern language models stems from two critical innovations: sub-word tokenization and compute-optimal scaling. Transitioning from character-level to Byte-Pair Encoding~\cite{sennrich2016neural} allowed models to compress text into tokens (word fragments), establishing an efficient underlying vocabulary. Similarly, research by \citet{hoffmann2022training} has shown that scaling in both the data and compute axes significantly decreases loss, leading to the development of extremely large models scaling on both axes, while \citet{kaplan2020scaling} provided a more compute-optimal training regime. 

Following a similar trajectory, Protein Language Models (PLMs) have shown strong performance at both protein sequence- and structure-level tasks. While the major focus of PLM development has been on scaling laws to boost performance, the role of tokenization granularity has received comparatively little attention in PLMs. Unlike human language, nature’s vocabulary that is protein domains, has evolved built-in mutation tolerance to preserve structural robustness. During evolution, structure is often preserved despite extensive sequence variation, and structural constraints play a central role in determining protein function.

The Ship of Theseus is a paradoxical question: if one by one all components of a ship are replaced, does the ship remain the same? While answering the paradox is tough, for folds it is answerable: the fold remains the same as long as amino acids in the sequence maintain its structural integrity. These amino acids are often found within evolutionarily conserved regions. Their presence enables the sequence to retain access to the same folding landscape. To understand it more clearly, biologists align a variety of sequences that result in the same fold into a 2D grid, a Multiple Sequence Alignment (MSA), revealing a landscape of conserved regions that are crucial for protein folding, which in turn governs its function. Since structure is more conserved than sequence\cite{illergaard2009structure}, the sequence-to-structure mapping is a many-to-one problem. To address this, researchers have developed various substitution matrices\cite{henikoff1992amino} that define the penalty of mutability between a pair of amino acids using empirical data. 

Structural classification systems reveal that proteins with highly divergent sequences can share common folds and the task of detecting them is known as remote homology inference\cite{rost1999twilight}. In these systems\cite{orengo1997cath,cheng2014ecod}, a fold is a conceptual entity that represents all individual domains within a given level of the classification hierarchy. The classification levels can have different arrangements and names, ranging from finest to coarse grain, such as CATH~\cite{orengo1997cath}:Class, Architecture, Topology, Homology which differs from SCOP~\cite{brenner199637} annotations Class, Fold, Superfamily, Family. The Class-level contains collections of alpha elements (spiral coil shape), beta elements (flat sheet), or a combination of them. The divergence of these sequences can be computed as the number of amino acid differences between aligned sequences. Applying a threshold of 30\% sequence identity as a definition of remote homology, \citet{kabir2024twilight} demonstrates that PLMs still fail to detect remote homologs and assign them correctly within structural classification hierarchies. Remote homology inference is, thus, considered one of the hardest tasks in this field. 

Structural information greatly simplifies homology inference. Although structure-prediction models have made incredible progress, protein folding problem, the essential rules governing fold conservation and the problem of function prediction remain unsolved \cite{jumper2021highly,abramson2024accurate,doi:10.1126/science.ads0018}. If the sequence holds the fundamental ground truth of function, it is inherently more valuable to decode this from sequence space rather than treating the two modalities as isolated spaces. 

If fold survival constrains sequence divergence, conserved MSA zones can act as the fundamental vocabulary of nature. In this work, we demonstrate that incorporating knowledge from the biological domain directly into the tokenization stage can substantially improve protein language model efficiency and performance without scaling in either axis. Primarily our contributions are:
\begin{itemize}
    \item \textbf{Zoned Encoding of Sequence Traits (ZEST):} Creating a method to inject known domain priors as the vocabulary, thereby bypassing the need to relearn these priors during training and compressing the context far beyond any existing PLM.
    \item \textbf{Trie-Dropout:} A novel tokenization regularization technique that stochastically decomposes longer ZEST tokens into their constituent sub-zones during training, ensuring uniform vocabulary utilization and enabling Test Time Augmentation (TTA) at inference.
    \item \textbf{Layered Extraction of Molecular Ordering from Nature (LEMON):} A compact model trained on a single H100 GPU in one week that sets a new efficiency-accuracy Pareto frontier for  remote homology retrieval.
\end{itemize}

\begin{figure}[htbp]
    \centering
    \includegraphics[width=\textwidth]{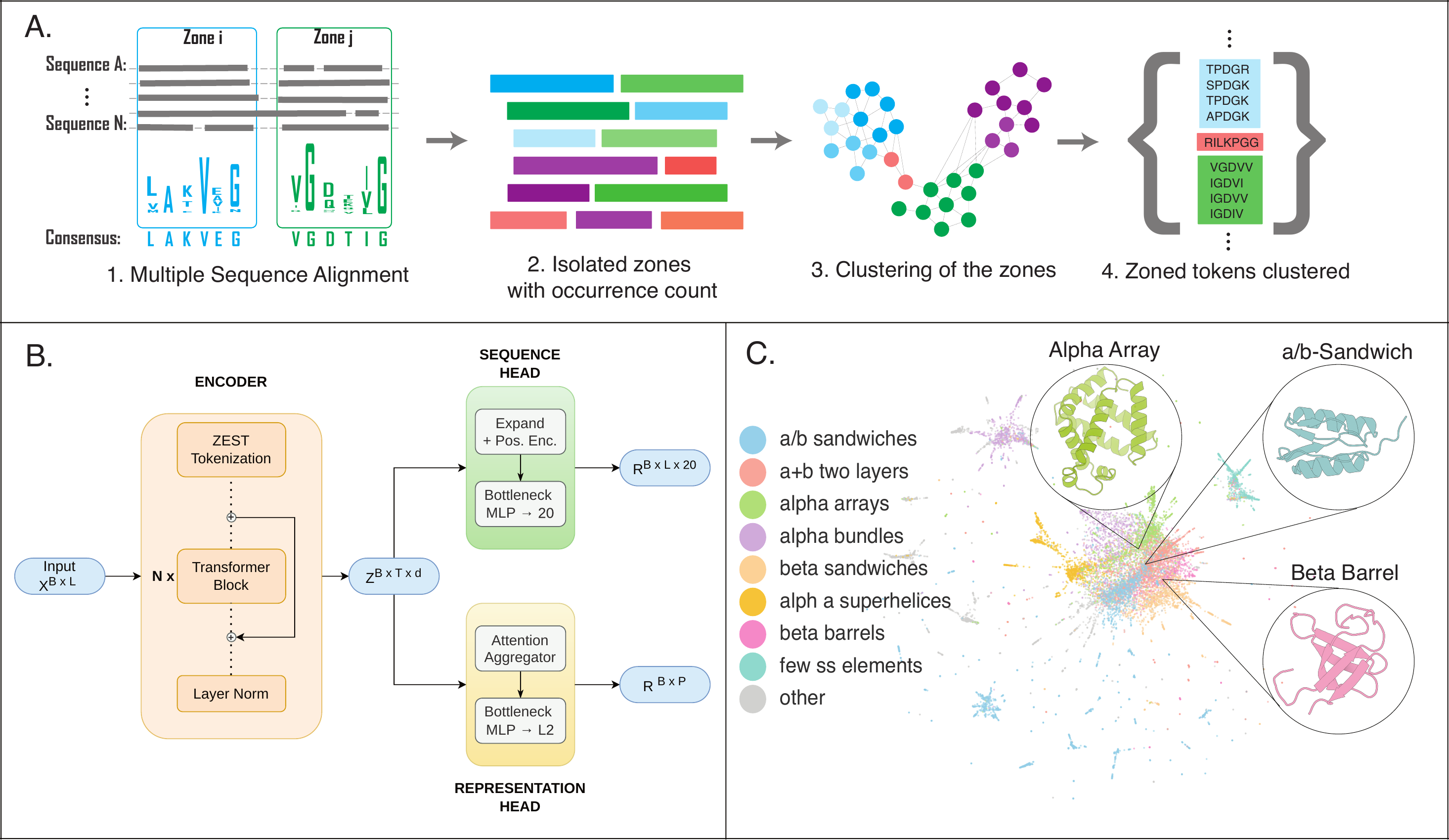}
    \caption{Overview of the proposed architecture and methodology. (A) The biological prior-injected vocabulary generation process. (B) The dual-head Y-architecture model design. (C) A reduced-dimensionality pacmap projection \cite{wang2021understanding} of the latent representations of protein domains, illustrating clear spatial separation between alpha helices, beta sheets, and their combinations.}
    \label{fig:architecture}
\end{figure}

\section{Related Works}
\textbf{Profile-Based Homology Inference.} Classical approaches to remote homology inference rely on profile-based methods built from MSAs. These profiles are based on Hidden Markov Models, considering a protein sequence as a 
Markovian Decision Process (MDP) over match, insert or delete state and residues being the emission. Tool-kits such as HMMER3~\cite{eddy2011accelerated}, HH-suite3~\cite{steinegger2019hh} construct these profiles from MSAs and use dynamic programming for similarity search. Despite their interpretability, these methods have the fundamental limitation of the Markovian Assumption (the future is conditionally independent of the past, given the present), which prevents them from capturing long range dependencies over the sequence governing fold. Curated collections of HMMs, such as Pfam~\cite{paysan2025pfam}, TIGRFAMs~\cite{haft2003tigrfams} and Gene3D~\cite{lewis2018gene3d}, have been paramount for identification and annotation of folds from sequence.  TEDLH~\cite{carreno2026tedlh} is a recent library of HMM profiles derived from The Encyclopedia of Domains (TED)~\cite{doi:10.1126/science.adq4946}. TED leverages structure-based domain segmentation tools to describe nearly 365 million domains in the alphafold database (AFDB)~\cite{varadi2022alphafold}.  

\textbf{Sequence-Based Protein Language Models.} Recent advances in protein language models have leveraged transformer~\cite{vaswani2017attention} architectures to learn contextual representations directly from raw protein sequences with the ability of bi-directional attention. The field pivoted to pre-train sequence based PLMs with BERT-style Masked Language Modeling (MLM)~\cite{devlin2019bert} on larger sequence corpora from UniRef~\cite{suzek2007uniref}. Models trained in this fashion demonstrate that raw PLM embeddings capture biophysical properties of the sequence without any structural supervision. Taking a different approach, MSAtransformers~\cite{rao2021msa} operate on aligned sequences using axial attention across rows and columns allowing them to capture co-evolutionary patterns. Despite impressive performance, residue based models are still limited by the sequence noise limiting their capacity~\cite{kabir2024twilight}.

\textbf{Contrastive Learning For Remote Homology.} To address the limitations of masked language modeling for similarity search, recent work has incorporated contrastive learning objectives to better align protein embeddings with structural relationships. The standard MLM training optimizes denoising of the sequence rather than similarity search. Models such as ProtTucker~\cite{heinzinger2022contrastive} address this by applying contrastive learning to PLM embeddings to optimize PLM embeddings for CATH hierarchy towards structural similarity.  Other research groups take a more direct approach, aligning sequence model embeddings to structure-trained PLM embeddings to incorporate structural supervision into sequence-only models~\cite{https://doi.org/10.1002/advs.202404212}.

\textbf{Structure-Aware Vocabulary.} Several recent approaches attempt to incorporate structural information into protein language models through structure-aware tokenization or joint sequence–structure representations. Thus, models such as SaProt~\cite{su2023saprot} introduce structure aware vocabulary including 3Di structure derived tokens from Foldseek~\cite{van2024fast}. SaProt achieves strong performance utilizing these tokens but requires explicit structure as input limiting it's capacity for protein with no known structures. ProstT5~\cite{heinzinger2024sty} takes a bilingual approach of treating these 1D sequence and 3D structure as two languages. More recently, work on protein structure tokenization has explored encoding local 3D context into discrete representations to improve structure informed language modeling. While GeoBPE~\cite{sun2025protein} proposed applying BPE to geometric protein structures. All these methods have similar dependency of structure availability at inference time, limiting the real world applications. CATHe~\cite{nallapareddy2023cathe} trained an artificial neural network directly on ProstT5 embeddings to classify sequences into CATH superfamilies, targeting sequences below 20\% (CATHS20) sequence identity.

\textbf{Tokenization Strategies.} PUMA~\cite{suyunu2025puma} challenged this by augmenting the standard BPE algorithm with evolution aware mutations, using substitution matrices to generate candidate token pairs, but remains theoretically proposed without trained models and relies on statistical mutation tendencies rather than empirically observed conservation.

\section{Methods}
\subsection{Vocabulary}
To construct a biologically informed vocabulary, we identified conserved and synonymous multi-residue patterns, zones($\ge 3$ residues \& is not homo-polymer) from HHsuite's consensus sequences for 765,248 HMM profiles (TEDLH)~\cite{carreno2026tedlh}. These zones are further clustered using MMseqs2~\texttt{linclust}~\cite{steinegger2017mmseqs2} at 70~\% sequence identity. To prioritize biologically significant motifs, each cluster was ranked using a score that aggregates cross-profile frequency while prioritizing zone length
From this ranking, we select top 31,975 clusters.%(see Appendix~\ref{app:zest}). 
To formalize the vocabulary, we append 20 standard amino acids as residue level fallback tokens and 5 special tokens. Each member of the clusters are assigned the same token id, inherently addressing the many-to-one problem. The resulting vocabulary \textbf{Zoned Encoding of Sequence Traits (ZEST)}, achieves a mean token length of mean length $\sim4$ residues, yielding meaningful context compression. 

%\footnote{$score = freq \cdot length^{1.5}$ --- ensuring longer motifs are not penalized by their naturally lower raw frequencies.}.

\begin{wrapfigure}{r}{0.5\textwidth}
    \centering
    \vspace{-10pt}
    \includegraphics[width=0.45\textwidth]{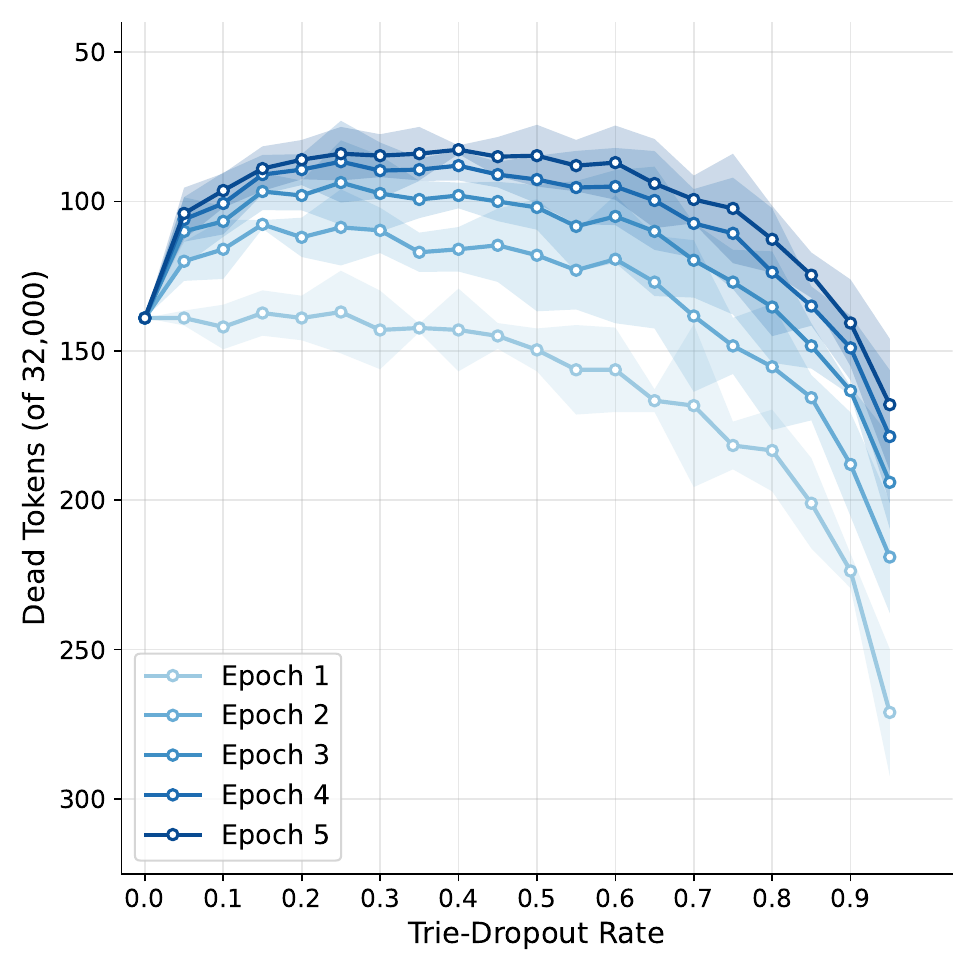}
    \caption{Multi-seed analysis of vocabulary coverage across one million sequences across epochs.}
    \label{fig:dropout}
\end{wrapfigure}
% \includegraphics[width=\textwidth]{figures/dropout_effect.pdf}
% \caption{Multi-seed analysis of vocabulary coverage across one million sequences spanning 5 epochs.}

Protein evolution reuses sequence over different length ranges: a long conserved motif often contains shorter conserved sub-motifs as prefixes. Meaning that shorter sub-patterns---which may themselves carry distinct structural semantics---are never exposed to the encoder as independent tokens. Figure~\ref{fig:dropout} shows that nearly 139 tokens are left completely unattended after visiting 5 millions sequences due to greedy longest-match tokenizer will always consume the longest available zone at each position. An analogous problem exists in sub-word NLP tokenization, where BPE-dropout~\cite{provilkov2020bpe} addresses it by randomly omitting merge operations; however, BPE merges are learned from corpus statistics and the resulting segmentation has no guaranteed structural correspondence. To mitigate this issue, we built a prefix tree (trie); where a vocabulary entry spanning $k$ residues naturally nests within longer entries that share the same prefix, providing a biologically grounded hierarchy of segmentation granularity.

% \begin{wrapfigure}{r}{0.5\textwidth} % {l} for left, 40% of text width
%     \centering
%     \includegraphics[width=0.45\textwidth]{figures/tta_sweep.pdf}
%     \caption{Something}
%     \label{fig:tta}
% \end{wrapfigure}
During tokenization, the trie matcher at each position returns \emph{all} valid matches from longest to shortest. With probability $p_{\text{drop}}$, the greedy choice is overridden and a uniformly sampled shorter match is selected instead, forcing the encoder to process the remaining suffix as one or more separate tokens. Serving as a augmentation, the dropout ensures the same protein sequence is segmented differently across iterations, preventing the model from relying solely on surface-level long-zoned identity. At the highest dropout the vocab falls back to individual characters. Concretely, for a position where the trie yields matches of lengths $(k_1 > k_2 > \cdots > k_m)$, trie dropout selects $k_j$ ($j > 1$) with probability $p_{\text{drop}} / (m{-}1)$ per alternative, otherwise retaining the greedy choice $k_1$. 
% We apply trie dropout at $p_{\text{drop}} = 0.15$ during our entire training phases. 

% \begin{itemize}
% \item We go through 765K such Profiles from TED-LH and define put vocab by taking all conserved islands
%     \subitem islands: appear more than once across profiles and is not water 
%     \subitem water: variable/non-conserver region 
%     \subitem Then we cluster them based on BLOSUM using MMSEQ2 (Cit'em)
%     \subitem We take top 31975 top clusters + 20 AAs + 5 special tokens as vocab
%     \subitem to address this many to one problem we assign same token id to all members of the cluster
%     \subitem Hence ZEST

% \item Intro To Trie dropout
%     \item Nature likes to reuse successful patterns, that same is evident in the islands. 
%     \item A greedy matching would mean the the sub token never gets attended
%     \item Possible solution would be BPE-Dropout that breaks randomly but for natural language that may make sense, due to repeating patterns we can build a prefix tree
%     \item The dropout basically forces the model to explore the children instead of just relying on the surface level
    
% \item Figure 1: 3-panel tokens, the Trie, abliteration to BPE, and compression

% \end{itemize}

\subsection{Model Architecture}

% To capture both local residue-level features and global representation properties, we design a Y-shaped transformer architecture with dual heads in Figure~\ref{fig:architecture}, to mitigate high variability of protein sequence lengths. With the help of dual head gradient flow for the multi task learning, the encoder learns rich yet generalizable features. 

% \paragraph{Motivation.}
Standard protein language models operate at the single-residue level, processing sequences of length $L$ with $O(L^2)$ attention cost.
The ZEST tokenizer replaces this with a 32K-entry vocabulary greedy max-match, compressing protein sequences by 4--6$\times$.
This compression is not merely computational; each multi-residue token corresponds to a recurrent characteristic signature, injecting domain-aware inductive bias directly at the input representation. 
% \paragraph{Compression and the reconstruction problem.}

Tokenization compresses a sequence of $L$ amino acids into $T \ll L$ tokens, where each token spans a variable number of residues recorded in a \texttt{token\_lengths} vector.
This creates a fundamental tension: the compressed token-level representation is efficient for global reasoning (attention, pooling), but downstream tasks such as residue-level properties require recovering per-residue embeddings. LEMON resolves this through a shared encoder whose token embeddings are simultaneously consumed by two task-specific heads Figure~\ref{fig:architecture}~B.% and in Appendix~\ref{app:dual-head} we show how the sequence head helps preserve isotropy.

\paragraph{Shared Encoder.}
The encoder is a pre-norm transformer stack with Rotary Position Embeddings (RoPE)~\cite{su2024roformer}, SwiGLU feed-forward networks~\cite{shazeer2020glu}, and Flash Attention~\cite{dao2022flashattention} via PyTorch's \texttt{scaled\_dot\_product\_attention}. Each attention block applies pre-LayerNorm, projects to queries/keys/values (bias-free), applies RoPE to Q and K, computes attention, then feeds through a gated FFN: $\text{FFN}(x) = W_3\bigl(\text{SiLU}(W_1 x) \odot W_2 x\bigr)$, with hidden dimension $\lfloor \tfrac{2}{3} d \cdot m \rfloor$ where $m$ is the FFN multiplier. Padding tokens are zeroed in keys and values before attention and in the output after each sub-layer. Token embeddings are tied to the MLM output projection, i.e.\ the token-level MLM logits are computed as $\text{logits} = z \cdot W_{\text{emb}}^\top$.

\textbf{Representation head.}
For hierarchical contrastive learning, variable-length token embeddings are projected into a fixed-size sequence representation.
We implement attention pooling that uses a single learned query vector $q_0 \in \mathbb{R}^{d}$ (initialized as $\mathcal{N}(0, 0.02)$) attending over all token embeddings via 4-head scaled dot-product attention for projecting $z \in \mathbb{R}^{B \times T \times d}$ :
\begin{equation}
f_{\text{seq}} = \text{LayerNorm}\!\bigl(W_O \cdot \text{Attn}(W_Q q_0,\; W_K z,\; W_V z)\bigr),
\end{equation}
followed by a learned alignment projection initialized to identity.
The resulting $f_{\text{seq}} \in \mathbb{R}^d$ is then passed through a bottleneck projector---a stack of $N_{\text{proj}}$ blocks, each containing two sub-layers forming a $d{\to}h{\to}d$ bottleneck---followed by a final linear projection to the contrastive embedding dimension $d_{\text{proj}}$ and $\ell_2$-normalized.
% The output is  and scaled by a learned temperature $\tau = \exp(\log \tau_0)$, clamped to $[0.01, 1.0]$.
% For the LEMON model: $d{=}768$, $h{=}3072$ ($4\times$), $N_{\text{proj}}{=}4$ blocks, $d_{\text{proj}}{=}384$.
% $[\text{Linear}(d{\to}h), \text{LN}, \text{GELU}, \text{Dropout}]$

\textbf{Sequence head.} The sequence head reconstructs per-residue amino acid predictions from the encoder's compressed token representations, effectively inverting the tokenization. Given encoder outputs $z \in \mathbb{R}^{B \times T \times d}$, each token embedding is first broadcast across its residue span to produce $\tilde{z} \in \mathbb{R}^{B \times L \times d}$. Each residue then receives two additive positional signals: a global embedding $p_g$ encoding its absolute position in the chain, and a local embedding $p_\ell$ (ALiBi-style \cite{press2021train}) encoding its offset within the token span. The combined representation $\hat{z} = \tilde{z} + p_g + p_\ell$ is passed through a two-block bottleneck MLP and a final linear projection to 20-way amino acid logits.

The dual positional encoding serves a specific purpose: the global signal prevents the model from conflating the same conserved motif appearing at different chain positions, while the local signal preserves sub-token residue identity. Together, the MLM gradient from this head penalizes any loss of fine-grained sequence information during encoding, keeping the shared encoder honest about local residue content despite operating on coarse token boundaries.

This design enables the shared encoder to maintain residue-level fidelity despite operating on compressed tokens: the expansion head's gradient signal (via residue MLM) penalizes any loss of local sequence information during encoding, while the global positional encoding prevents the model from confusing identical motifs occurring at different chain positions.

% \begin{itemize}
%     \item Motivation:
    
%     \item Talk about compression
%     \item Shared Encoder
%     \item Sequence head Expansion logic 
%     \item Representation Head Attention Pooling
% \end{itemize}

% \begin{figure}[ht]
%     \centering
%     \includegraphics[width=\textwidth]{figures/lemon_design.png}
%     \caption{Architecture of the LEMON: The sequences are embedded and passed through 24 transformer blocks with Rotary Position Embedding. The architecture follows a dual head Y-Network design with shared encoder. The shared representation $\mathbf{z} \in \mathbb{R}^{B \times T \times D}$. The Sequence Head expands \textit{T} tokens back to \textit{L} characters by replicating the tokens across characters(residues) and global token level \& local character level learned positional encoding. While the Representation Head pools via learnable query attention and projects to a contrastive embedding space with learned temperature $\tau$.}
%     \label{fig:architecture}
% \end{figure}

\subsection{Data}
We pre-train on 50\% of UniRef90~\cite{suzek2015uniref}, containing $\sim$90M protein sequences with the objective of token level and sequence level MLM and fine-tuning on 59.7M protein pairs from TEDLH domain annotations, each labeled at three levels of the CATH~\cite{orengo1997cath} hierarchy: Architecture, Topology, and Superfamily. The model was trained on sequences alone and contrastive learning was applied at the hierarchy level. Training and validation pairs are separated by CD-HIT\cite{li2006cd} to prevent sequence level leakage within the finetuning set. For evaluations, we rely on standard SCOPe~\cite{chandonia2019scope} and CATH~S20.

We conducted a rigorous audit between finetuning sequences (726,922 TED domains) and all the benchmarks. Exact string matching found zero identical sequence. An all-against-all search using cd-hit-2d revealed that $\sim10\%$ of benchmark domains share $\ge 70\%$ global sequence identity with a training sequence, and $\sim2.5\%$ share $\ge90\%$ identity---consistent with the natural redundancy of protein sequence space and well below the identity threshold at which benchmarks are evaluated. At the fold level, 80.9\% of $S20$ fold types appear in training, which is expected: remote homology benchmarks test sequence-distant members of known folds, not unseen fold classes. Collectively, these results confirm that benchmark performance reflects generalization across sequence divergence, not memorization of training sequences.

% \subsection{Data}
% \begin{itemize}
% \item Talk about the data
%     \subitem Uniref 90 with MLM and why 
%     \subitem Ted-LH and why is ideal 
%     \subitem some nerdy data statics  
%     \subitem Also exmplain

% \item PreTraining on Uniref90 50\% 
%     \subitem Dropout 15\%
%     \subitem Token MLM + Sequence MLM with 30\% masking 
%         \subitem Proteins sequences are water, 30\% masking helps here 
%     \subitem UncertaintyWeightedLoss
%     \subitem The pretraining results what ppl we achived on each task while the task is much harder 
    
% \item FineTuning on TED-LH CATH Classification 
%     \subitem 15\% masking and 15\% dropout
%     \subitem UncertaintyWeightedLoss combine together two heads
%     \subitem Hierarchical Contrastive Learning For sequence head
%     \subitem Sequence head on MLM-BERT style 

% \item Dual Pull of Y acting as regularization for embedding quality
    
% \end{itemize}

\section{Experiments}
\subsection{Remote Homology with LEMON}
\textbf{Task Definition:} We evaluate LEMON on the task of remote homology detection, where the goal is to retrieve proteins sharing the same fold as a query despite low sequence similarity. Each protein sequence is encoded into a single embedding vector, and retrieval is performed by ranking all candidate sequences by cosine similarity to the query. Using the SCOP and CATH classifications as ground truth, pairs are labeled positive if they agree at a given hierarchical level (e.g. SCOP Family and CATH Superfamily) but disagree at the level below, and negative otherwise~\cite{chen2018comprehensive, strodthoff2020udsmprot, rives2021biological}. This approach ensures that only remote homologs constitute positive pairs, making the benchmark a stringent test of learned protein representations.

\textbf{Evaluation baselines:}
With LEMON being finetuned on CATH hierarchy, our evaluations are done on both annotation schemes SCOP and CATH. Following evaluation criteria defined by \citet{kabir2024twilight}, we segment across sequence-identity thresholds(10–95\%) for SCOPe sequences. These evaluation benchmarks are constructed to ensure evaluation schema differs from training schema to ensure reliability. 

We compare LEMON against baselines spanning four families. \textit{Sequence PLMs (no contrastive supervision):} ESM-2~\citep{doi:10.1126/science.ade2574}, ESM3-open~\citep{doi:10.1126/science.ads0018}, and ProtTrans-T5-XL~\citep{elnaggar2021prottrans}, using mean-pooled embeddings and public weights. \textit{Structure-conditioned and multimodal PLMs:} SaProt~\citep{su2023saprot}, ProstT5~\citep{heinzinger2024sty}, DPLM-2.1~\citep{hsieh2025elucidating} and ProTrek~\citep{su2025trimodal}, which use structure tokens or additional modalities. \textit{Contrastive/retrieval models:} ProtTucker~\citep{heinzinger2022contrastive} finetunes ProtT5-XL with a CATH-supervised contrastive head, DHR~\citep{hong2025fast}, and TM-Vec~\citep{hamamsy2024protein}, which is trained with structural (TM-score) supervision. \textit{Matched controls:} LEMON-BPE (same architecture, data and finetuning, with BPE replacing ZEST) and ESM2-CFT (ESM-2 650M finetuned end-to-end with LEMON's head and data). All baseline embeddings were generated by us under identical conditions. We also include the HMM-profile method HHblits~\citep{steinegger2019hh}.

\textbf{Metrics:}
The difficulty of fold level retrieval arises in part due to the extreme class imbalance($\sim200:1$ negative-to-positive ratio on SCOP). While such imbalance is grounds for favoring Area Under Precision-Recall Curve~(AUPRC), \citet{mcdermott2024closer} show that class imbalance alone doesn't justify AUPRC over Area Under the Receiver Operating Characteristic Curve~(AUROC). AUROC treats all misranked pairs uniformly, whereas AUPRC disproportionately penalizes errors among high-scoring samples. Therefore, we adopt fold-level AUROC as our primary metric and offer mAP as a complementary measure of top-of-list retrieval quality.

\textbf{Results:}
Figure~\ref{fig:threholds}a depicts fold-level retrieval on SCOPe database across six sequence-identity thresholds, where lower thresholds retain only distant homologs making retrieval strictly harder. LEMON leads all sequence-only models at every threshold, from the most challenging (th10: 0.91) to the easiest (th95: 0.98), closely matching TM-Vec. Structure supervised SaProt trails LEMON by a large margin, underscoring that structural supervision at the residue level does not substitute for the global representation geometry learned by LEMON. 
Figure~\ref{fig:threholds}b shows the parameter efficiency compared to performance. HHblits outperforms all models that have not undergone contrastive fine-tuning, highlighting the extent to which standard PLM embeddings remain poorly calibrated for similarity search without explicit metric learning.  

\begin{table*}[tbp]
\centering
\caption{Complete benchmark results across CATH-S20 and SCOPe$@10\%$ classification levels. }
\label{tab:full_benchmark}

\resizebox{\textwidth}{!}{
\footnotesize

\begin{tabular}{l cccccccc}
\toprule
& \multicolumn{2}{c}{\textbf{CATH-S20 Architecture}} & \multicolumn{2}{c}{\textbf{CATH-S20 Topology}} & \multicolumn{2}{c}{\textbf{SCOPe Fold}} & \multicolumn{2}{c}{\textbf{SCOPe Superfamily}} \\
\cmidrule(lr){2-3} \cmidrule(lr){4-5} \cmidrule(lr){6-7} \cmidrule(lr){8-9}
\textbf{Model} & \textbf{AUROC} & \textbf{mAP} & \textbf{AUROC} & \textbf{mAP} & \textbf{AUROC} & \textbf{mAP} & \textbf{AUROC} & \textbf{mAP} \\
\midrule
LEMON 
& \wildcard[5.6em]{0.815}{$\pm$}{0.002} 
& \wildcard[5.6em]{0.346}{$\pm$}{0.003}
& \wildcard[5.6em]{0.886}{$\pm$}{0.002} 
& \wildcard[5.6em]{0.312}{$\pm$}{0.006} 
& \wildcard[5.6em]{0.901}{$\pm$}{0.004} 
& \wildcard[5.6em]{0.300}{$\pm$}{0.009} 
& \wildcard[5.6em]{0.942}{$\pm$}{0.003} 
& \wildcard[5.6em]{0.463}{$\pm$}{0.010} \\

TM-Vec 
& \wildcard[5.6em]{0.716}{$\pm$}{0.002} 
& \wildcard[5.6em]{0.236}{$\pm$}{0.003} 
& \wildcard[5.6em]{0.895}{$\pm$}{0.002} 
& \wildcard[5.6em]{0.347}{$\pm$}{0.005} 
& \wildcard[5.6em]{0.936}{$\pm$}{0.004} 
& \wildcard[5.6em]{0.448}{$\pm$}{0.011} 
& \wildcard[5.6em]{0.980}{$\pm$}{0.002} 
& \wildcard[5.6em]{0.677}{$\pm$}{0.009} \\

ProtTucker 
& \wildcard[5.6em]{0.794}{$\pm$}{0.002} 
& \wildcard[5.6em]{0.260}{$\pm$}{0.003} 
& \wildcard[5.6em]{0.881}{$\pm$}{0.002} 
& \wildcard[5.6em]{0.299}{$\pm$}{0.005} 
& \wildcard[5.6em]{0.873}{$\pm$}{0.004} 
& \wildcard[5.6em]{0.234}{$\pm$}{0.009} 
& \wildcard[5.6em]{0.961}{$\pm$}{0.002} 
& \wildcard[5.6em]{0.601}{$\pm$}{0.009} \\

LEMON-BPE (control) 
& \wildcard[5.6em]{0.710}{$\pm$}{0.002} 
& \wildcard[5.6em]{0.188}{$\pm$}{0.002} 
& \wildcard[5.6em]{0.845}{$\pm$}{0.002} 
& \wildcard[5.6em]{0.173}{$\pm$}{0.004} 
& \wildcard[5.6em]{0.865}{$\pm$}{0.005} 
& \wildcard[5.6em]{0.191}{$\pm$}{0.007} 
& \wildcard[5.6em]{0.931}{$\pm$}{0.003} 
& \wildcard[5.6em]{0.336}{$\pm$}{0.010} \\

ESM2-CFT (650M) 
& \wildcard[5.6em]{0.668}{$\pm$}{0.002} 
& \wildcard[5.6em]{0.175}{$\pm$}{0.002} 
& \wildcard[5.6em]{0.829}{$\pm$}{0.002} 
& \wildcard[5.6em]{0.215}{$\pm$}{0.004} 
& \wildcard[5.6em]{0.870}{$\pm$}{0.004} 
& \wildcard[5.6em]{0.251}{$\pm$}{0.008} 
& \wildcard[5.6em]{0.970}{$\pm$}{0.002} 
& \wildcard[5.6em]{0.655}{$\pm$}{0.009} \\

DHR 
& \wildcard[5.6em]{0.596}{$\pm$}{0.002} 
& \wildcard[5.6em]{0.124}{$\pm$}{0.001} 
& \wildcard[5.6em]{0.715}{$\pm$}{0.003} 
& \wildcard[5.6em]{0.111}{$\pm$}{0.003} 
& \wildcard[5.6em]{0.752}{$\pm$}{0.006} 
& \wildcard[5.6em]{0.125}{$\pm$}{0.006} 
& \wildcard[5.6em]{0.933}{$\pm$}{0.003} 
& \wildcard[5.6em]{0.473}{$\pm$}{0.010} \\

ESM3-open (1.4B) 
& \wildcard[5.6em]{0.591}{$\pm$}{0.002} 
& \wildcard[5.6em]{0.137}{$\pm$}{0.001} 
& \wildcard[5.6em]{0.677}{$\pm$}{0.003} 
& \wildcard[5.6em]{0.097}{$\pm$}{0.002} 
& \wildcard[5.6em]{0.712}{$\pm$}{0.006} 
& \wildcard[5.6em]{0.115}{$\pm$}{0.005} 
& \wildcard[5.6em]{0.857}{$\pm$}{0.004} 
& \wildcard[5.6em]{0.340}{$\pm$}{0.008} \\

ProTrek (650M) 
& \wildcard[5.6em]{0.586}{$\pm$}{0.002} 
& \wildcard[5.6em]{0.118}{$\pm$}{0.001} 
& \wildcard[5.6em]{0.692}{$\pm$}{0.003} 
& \wildcard[5.6em]{0.066}{$\pm$}{0.002} 
& \wildcard[5.6em]{0.709}{$\pm$}{0.005} 
& \wildcard[5.6em]{0.073}{$\pm$}{0.004} 
& \wildcard[5.6em]{0.874}{$\pm$}{0.004} 
& \wildcard[5.6em]{0.261}{$\pm$}{0.008} \\

DPLM-2.1 (650M) 
& \wildcard[5.6em]{0.577}{$\pm$}{0.002} 
& \wildcard[5.6em]{0.128}{$\pm$}{0.001} 
& \wildcard[5.6em]{0.674}{$\pm$}{0.003} 
& \wildcard[5.6em]{0.101}{$\pm$}{0.002} 
& \wildcard[5.6em]{0.698}{$\pm$}{0.006} 
& \wildcard[5.6em]{0.114}{$\pm$}{0.005} 
& \wildcard[5.6em]{0.829}{$\pm$}{0.005} 
& \wildcard[5.6em]{0.316}{$\pm$}{0.009} \\

% DPLM-2 (650M)
% & \wildcard[5.6em]{0.570}{$\pm$}{0.002}
% & \wildcard[5.6em]{0.122}{$\pm$}{0.001}
% & \wildcard[5.6em]{0.642}{$\pm$}{0.003}
% & \wildcard[5.6em]{0.080}{$\pm$}{0.002}
% & \wildcard[5.6em]{0.671}{$\pm$}{0.005}
% & \wildcard[5.6em]{0.082}{$\pm$}{0.004}
% & \wildcard[5.6em]{0.815}{$\pm$}{0.005}
% & \wildcard[5.6em]{0.258}{$\pm$}{0.008} \\

\bottomrule
\end{tabular}%
}
\end{table*}

Further shown in Table~\ref{tab:full_benchmark}, LEMON achieves the best CATH-S20 Architecture scores and being competitive with TM-Vec. Despite LEMON being 200M parameters and sequence only model, it is extremely competitive and sometimes outperforms all SOTA PLMs, and the matched BPE, EMS2-CFT control, which indicates that tokenization contributes beyond the contrastive objective. Indicating that representation quality drives performance rather than brute-force scale and contrastive metric learning on sequence tokens alone suffices for fold-level discrimination. %We discuss steps for reproduction in Appendix~\ref{app:reproduction}.

% \begin{itemize}

% \item A brief description of each databases SCOPE SCOP CATHs20
% \item definition of the setup of testing conditions
% \item One big table of detailed task specific benchmarks
% \item Aggrigated Table of ROC across thresholds  
% \item Figure on ROC vs Parameter efficiency
    
% \end{itemize}

% \input{bodies/tables/zest_ablt}
\subsection{Test Time Augmentation:}

\begin{wrapfigure}{l}{0.5\textwidth}
    \vspace{-18pt}
    \centering
    \includegraphics[width=0.48\textwidth]{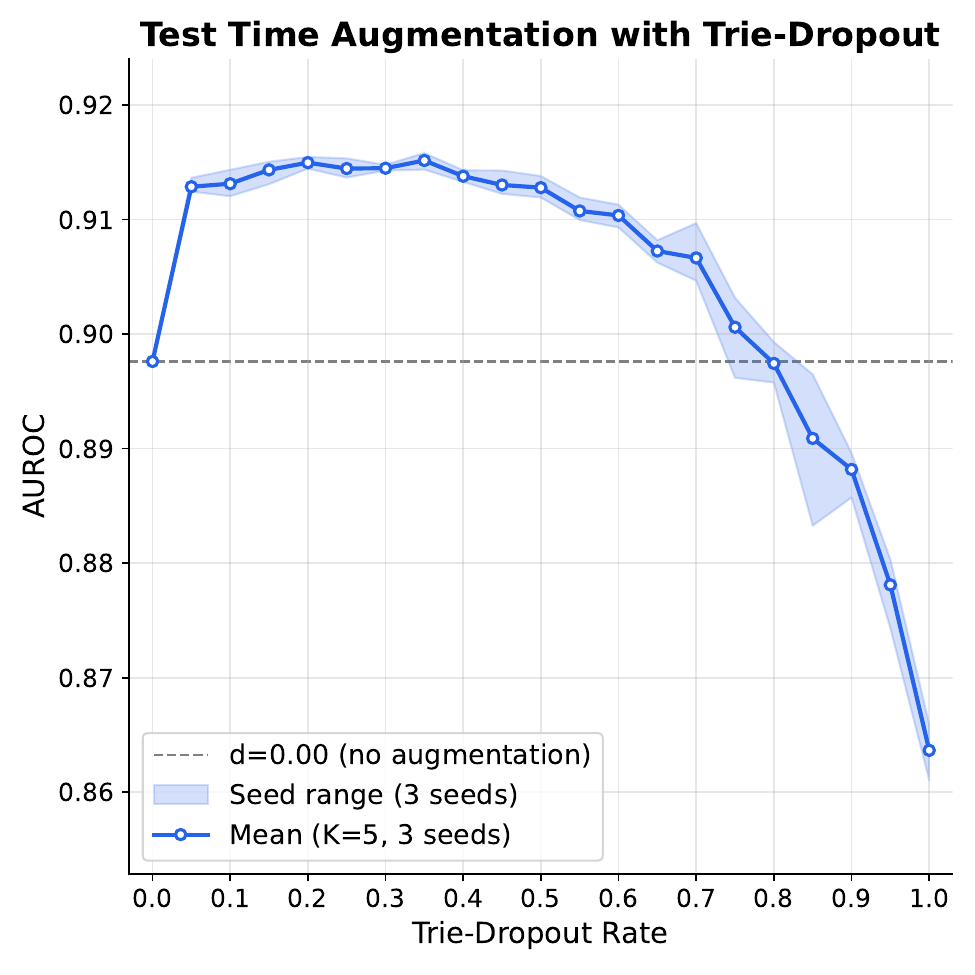}
    \caption{\small Multi-seed test time augmentation effect.}
    \label{fig:tta}
    \vspace{-18pt}
\end{wrapfigure}

Because ZEST's trie-based tokenizer supports stochastic segmentation, it naturally enables test-time augmentation (TTA) at inference by generating $K$ stochastic tokenizations of each sequence and averaging the resulting embeddings,
\begin{equation*}
\bar{\mathbf{e}} = \frac{1}{K} \sum_{k=1}^{K} f_\theta\!\bigl(\textsc{Tokenize}(\mathbf{x};\, d)\bigr)
\end{equation*}
where $d$ is the trie-dropout rate and $f_\theta$ is frozen LEMON. We sweep $d \in [0, 1]$ with $K=5$ passes per sequence, evaluating in SCOP fold-level retrieval across three random seeds; results are shown in Figure~\ref{fig:tta}.

The resulting inverted-U profile has two implications. First, moderate trie-dropout provides a reliable, training-free improvement to retrieval quality: embeddings from multiple passes are averaged, so the computational overhead scales linearly with $K$ and carries no memory cost beyond a single forward pass. Second, the breadth of the plateau, spanning roughly $d \in [0.05, 0.5]$, means that practitioners need not tune $d$ precisely; any value in this range recovers the majority of the available gain, making the method robust to hyperparameter choice in deployment.

\subsection{Computational efficiency}
We benchmark inference cost on a single L40S GPU across three length bins: small (100-499 residues), medium (500-999) and large (1000-1999) of protein residue lengths. ESM2 handles at most 1,022 residues, so the large bin is evaluated only for LEMON and ProstT5. Table~\ref{tab:speedup_zest} shows pattern is consistent: LEMON is the fastest and lightest model in every bin, and its advantage grows with length. On short sequences it is already several times faster than the ESM2 models, and on medium-length sequences the gap becomes an order of magnitude over ESM2-650M, with memory use staying nearly flat where the baselines’ grows steeply. Per-sequence compute shows the same trend. In the large bin, LEMON processes proteins that residue-level models cannot take in a single pass at roughly 2.8 ms, whereas ProstT5, the only baseline able to run there, is more than an order of magnitude slower and needs about ten times the memory. These gains come from two sources: ZEST shortens the input by roughly 3.3x, and LEMON’s encoder is smaller than the larger baselines. We treat the combined effect as the practically relevant figure, since compression is what makes the smaller encoder viable at this accuracy. Measurements cover the encoder forward pass only till residue level reconstruction.
\begin{table}[!t]
\centering
\caption{Model Performance Benchmarks by Sequence Length}
\label{tab:speedup_zest}
\small
\renewcommand{\arraystretch}{1}
\setlength{\tabcolsep}{8pt}
\begin{tabular}{@{}lccccc@{}}
\toprule
{\textbf{Model}} & {\textbf{Length Bins}} & \textbf{Mean(ms)} & \textbf{P95(ms)} & \textbf{Peak mem(GB)} & \textbf{GFLOPs} \\
\midrule

% LEMON
\multirow{3}{*}{\shortstack[l]{\textbf{LEMON}\\\footnotesize (204M)}} 
& \wildcard[5.2em]{100}{--}{499}   & \wildcard{$1.80$}{$\pm$}{$0.0$} & \wildcard{$1.90$}{$\pm$}{$0.0$} & \makebox[2.8em][r]{0.44} & \makebox[2.8em][c]{24} \\
& \wildcard[5.2em]{500}{--}{999}   & \wildcard{$2.00$}{$\pm$}{$0.0$} & \wildcard{$2.10$}{$\pm$}{$0.0$} & \makebox[2.8em][r]{0.47} & \makebox[2.8em][c]{73} \\
& \wildcard[5.2em]{1000}{--}{1999} & \wildcard{$2.80$}{$\pm$}{$0.0$} & \wildcard{$3.00$}{$\pm$}{$0.0$} & \makebox[2.8em][r]{0.53} & \makebox[2.8em][c]{159} \\
\midrule

% ESM2-150M
\multirow{3}{*}{\shortstack[l]{\textbf{ESM2}\\\footnotesize (150M)\rule[-3ex]{0pt}{0pt}}} 
& \wildcard[5.2em]{100}{--}{499}   & \wildcard{$4.00$}{$\pm$}{$0.1$}  & \wildcard{$5.60$}{$\pm$}{$0.1$}  & \makebox[2.8em][r]{0.49} & \makebox[2.8em][c]{72} \\
& \wildcard[5.2em]{500}{--}{999}   & \wildcard{$16.5$}{$\pm$}{$0.1$} & \wildcard{$18.6$}{$\pm$}{$0.1$} & \makebox[2.8em][r]{1.00} & \makebox[2.8em][c]{231} \\
% & \wildcard[5.2em]{1000}{--}{1999} & \makebox[2.2em][r]{6}    & \wildcard{$19.4$}{$\pm$}{$0.0$} & \wildcard{$19.4$}{$\pm$}{$0.0$} & \makebox[2.8em][r]{0.84} & \makebox[2.8em][r]{338} \\
\midrule

% ESM2-650M
\multirow{3}{*}{\shortstack[l]{\textbf{ESM2}\\\footnotesize (650M)\rule[-3ex]{0pt}{0pt}}} 
& \wildcard[5.2em]{100}{--}{499}   & \wildcard{$7.40$}{$\pm$}{$0.1$}  & \wildcard{$10.5$}{$\pm$}{$0.1$} & \makebox[2.8em][r]{1.51} & \makebox[2.8em][c]{311} \\
& \wildcard[5.2em]{500}{--}{999}   & \wildcard{$26.8$}{$\pm$}{$0.2$} & \wildcard{$29.9$}{$\pm$}{$0.1$} & \makebox[2.8em][r]{2.07} & \makebox[2.8em][c]{972} \\
% & \wildcard[5.2em]{1000}{--}{1999} & \makebox[2.2em][r]{6}    & \wildcard{$30.1$}{$\pm$}{$0.0$} & \wildcard{$30.1$}{$\pm$}{$0.0$} & \makebox[2.8em][r]{1.89} & \makebox[2.8em][r]{1400} \\
\midrule

% ProstT5
\multirow{3}{*}{\shortstack[l]{\textbf{ProstT5}\\\footnotesize (1.2B)}} 
& \wildcard[5.2em]{100}{--}{499}   & \wildcard{$13.8$}{$\pm$}{$0.2$} & \wildcard{$18.3$}{$\pm$}{$0.2$} & \makebox[2.8em][r]{3.16} & \makebox[2.8em][c]{431} \\
& \wildcard[5.2em]{500}{--}{999}   & \wildcard{$50.8$}{$\pm$}{$0.3$} & \wildcard{$57.5$}{$\pm$}{$0.1$} & \makebox[2.8em][r]{5.48} & \makebox[2.8em][c]{1350} \\
& \wildcard[5.2em]{1000}{--}{1999} & \wildcard{$60.5$}{$\pm$}{$0.0$} & \wildcard{$60.7$}{$\pm$}{$0.0$} & \makebox[2.8em][r]{5.63} & \makebox[2.8em][c]{2990} \\
\bottomrule
\end{tabular}
\end{table}

\subsection{Isolating effects of ZEST}
To justify ZEST vocabulary design and examine scaling behavior, we evaluate ZEST against standard character-level tokenization and a corpus-frequency BPE baseline across three random seeds and model scales (Table~\ref{tab:multiseed_scope}). At small scale, We pretrain 9 models with identical backbone architecture (50M encoder $+$ 8M head; $d=512$) and varying embedding tables (16.4M for ZEST \& BPE, 0.01M  for Char) on a stratified 20M subset of Uniref90 and then fine-tuned in the entire TED-LH dataset. At full scale, LEMON-BPE replicates the exact pre-training and fine-tuning pipeline of LEMON, while ESM2-CFT is end-to-end contrastive fine-tune of a pre-trained ESM-2 650M backbone using LEMON's representation head architecture.

The small-scale results highlights that while BPE accelerates pre-training throughput over character tokenization, its sharp decline in precision yields diminishing returns—demonstrating that naive NLP subword heuristics fail on protein sequences. In contrast, ZEST eliminates the trade-off by implicitly injecting the evolutionary conservation, it improves accuracy over both baselines while achieving the fastest fine-tuning time. Crucially, this performance gap widens at full scale: scaling capacity does not rescue LEMON-BPE. Moreover, despite operating at under one-third the parameters and compute of ESM2-CFT, LEMON outperforms it on remote fold retrieval where sequence divergence is highest. This underscores the efficiency and structural fidelity of ZEST across model and data scales.

% SCOPe multi-seed ablation (frozen finetune, best checkpoint).
% Three seeds per tokenizer: {42, 7, 123}. Reported as mean $\pm$ std.
% n_sequences = 7117, n_queries = 5350.
\begin{table}[!tb]
\centering
\caption{SCOPe structural retrieval benchmark on varying scale and tokenization.}
\label{tab:multiseed_scope}
\small
\setlength{\tabcolsep}{0pt}
\renewcommand{\arraystretch}{1}
\begin{tabular*}{\columnwidth}{@{\extracolsep{\fill}} lccccccc @{}}
\toprule
Tokenizer & Params & SF ROC $\uparrow$ & SF mAP $\uparrow$ & fold ROC $\uparrow$ & Fold mAP $\uparrow$ & $T_\text{pre}$ $\downarrow$ & $T_\text{ft}$ $\downarrow$ \\
\midrule
\multicolumn{8}{l}{\textit{Small scale training and finetuning}} \\
ZEST (32K) & 74M  & \wildcard[5.6em]{\textbf{0.739}}{\textbf{$\pm$}}{\textbf{0.017}} 
                  & \wildcard[5.6em]{\textbf{0.068}}{\textbf{$\pm$}}{\textbf{0.002}} 
                  & \wildcard[5.6em]{\textbf{0.749}}{\textbf{$\pm$}}{\textbf{0.014}} 
                  & \wildcard[5.6em]{\textbf{0.041}}{\textbf{$\pm$}}{\textbf{0.002}}
                  & 05h 04m & \textbf{16h 26m} \\
BPE (32K)  & 74M  & \wildcard[5.6em]{0.729}{$\pm$}{0.008} 
                  & \wildcard[5.6em]{0.036}{$\pm$}{0.002} 
                  & \wildcard[5.6em]{0.743}{$\pm$}{0.008} 
                  & \wildcard[5.6em]{0.021}{$\pm$}{0.001}
                  & \textbf{04h 38m} & 17h 10m \\
Char (25)  & 58M  & \wildcard[5.6em]{0.708}{$\pm$}{0.011} 
                  & \wildcard[5.6em]{0.059}{$\pm$}{0.002} 
                  & \wildcard[5.6em]{0.721}{$\pm$}{0.010} 
                  & \wildcard[5.6em]{0.036}{$\pm$}{0.002}
                  & 10h 50m & 17h 23m \\
\midrule
\multicolumn{8}{l}{\textit{Full scale training and finetuning}} \\
LEMON             & 200M & \wildcard[5.6em]{0.942}{$\pm$}{0.003} 
                         & \wildcard[5.6em]{0.463}{$\pm$}{0.010} 
                         & \textbf{\wildcard[5.6em]{0.901}{$\pm$}{0.004}} 
                         & \textbf{\wildcard[5.6em]{0.300}{$\pm$}{0.009}} 
                         & 122h & \textbf{46h 42m} \\
LEMON-BPE         & 200M & \wildcard[5.6em]{0.931}{$\pm$}{0.003} 
                         & \wildcard[5.6em]{0.336}{$\pm$}{0.010} 
                         & \wildcard[5.6em]{0.865}{$\pm$}{0.005} 
                         & \wildcard[5.6em]{0.191}{$\pm$}{0.007} 
                         & 110h & 48h 48m \\
ESM2-CFT          & 650M & \textbf{\wildcard[5.6em]{0.970}{$\pm$}{0.002}}
                         & \textbf{\wildcard[5.6em]{0.655}{$\pm$}{0.009}} 
                         & \wildcard[5.6em]{0.870}{$\pm$}{0.004} 
                         & \wildcard[5.6em]{0.251}{$\pm$}{0.008} 
                         & --- & 196h \\
\bottomrule
\end{tabular*}
\end{table}
% ($embed_{dim}$ = 512, layers = 16, $n_{head}$ = 8, $proj_{dim}$ = 128)

% Per-seed values (for reference / appendix):
%
% ZEST  seed 42:  ROC 0.7325  mAP 0.0679  fROC 0.7425  R@1 0.0839  R@5 0.1787  R@10 0.2407  R@20 0.3161
% ZEST  seed  7:  ROC 0.7576  mAP 0.0653  fROC 0.7643  R@1 0.0834  R@5 0.1785  R@10 0.2385  R@20 0.3161
% ZEST  seed 123: ROC 0.7262  mAP 0.0692  fROC 0.7393  R@1 0.0908  R@5 0.1847  R@10 0.2480  R@20 0.3232
%
% BPE   seed 42:  ROC 0.7373  mAP 0.0370  fROC 0.7510  R@1 0.0869  R@5 0.2114  R@10 0.2931  R@20 0.3964
% BPE   seed  7:  ROC 0.7301  mAP 0.0371  fROC 0.7434  R@1 0.0895  R@5 0.2150  R@10 0.2938  R@20 0.3929
% BPE   seed 123: ROC 0.7209  mAP 0.0345  fROC 0.7353  R@1 0.0858  R@5 0.1983  R@10 0.2684  R@20 0.3639
%
% Char  seed 42:  ROC 0.6960  mAP 0.0585  fROC 0.7099  R@1 0.0901  R@5 0.1895  R@10 0.2568  R@20 0.3357
% Char  seed  7:  ROC 0.7111  mAP 0.0585  fROC 0.7250  R@1 0.0847  R@5 0.1884  R@10 0.2568  R@20 0.3449
% Char  seed 123: ROC 0.7168  mAP 0.0612  fROC 0.7285  R@1 0.0886  R@5 0.1923  R@10 0.2615  R@20 0.3484

% \input{bodies/tables/fold_summary}
% \input{bodies/tables/zs_tasks}
\subsection{Zero-Shot Generalization}
% \begin{table}[H]
% \centering
% \resizebox{\linewidth}{!}{%
% \begin{tabular}{@{}l cc cc cc cc@{}}
% \toprule
% & \multicolumn{2}{c}{\textbf{Circ.\ Permutation}}
% & \multicolumn{2}{c}{\textbf{EC Number}}
% & \multicolumn{2}{c}{\textbf{GO MF}}
% & \multicolumn{2}{c}{\textbf{Domain Boundary}} \\
% \cmidrule(lr){2-3}\cmidrule(lr){4-5}\cmidrule(lr){6-7}\cmidrule(lr){8-9}
% & AUROC & AUPRC
% & F\textsubscript{max} & Prec.
% & F\textsubscript{max} & AUPR
% & NDO & NRes \\
% \midrule
% LEMON & \textbf{0.779} & \textbf{0.285}
%       & \textbf{0.821} & \textbf{0.865}
%       & 0.707 & 0.724
%       & \textbf{0.739} & 0.663 \\
% ESM-2 & 0.476 & 0.081
%       & 0.800 & 0.856
%       & 0.708 & 0.726
%       & 0.724 & \textbf{0.671} \\
% \bottomrule
% \end{tabular}
% }
% \caption{%
%     Zero-shot protein benchmarks (frozen embeddings, cosine $k$-NN, no task-specific training).
%     LEMON: 384-d projection, 200\,M params.
%     ESM-2: 1{,}280-d mean-pool, 650\,M params.%
% }
% \label{tab:zeroshot_main}
% \end{table}
\begin{table}[ht]
\centering
\caption{Zero-shot generalization across four protein tasks. All methods use frozen embeddings with cosine-similarity $k$-NN label transfer ($K\!=\!15$) or pairwise classification---no task-specific fine-tuning. The table shows that at a fraction of its parameter size, LEMON matches or exceeds SOTA zero-shot performance.}
\label{tab:zeroshot}

\footnotesize % Makes the text uniformly smaller without blowing up the table
\renewcommand{\arraystretch}{0.85} % Squeezes the vertical space between rows (default is 1.0)

\begin{tabular}{llccc}
\toprule
\textbf{Task} & \textbf{Metric} & \textbf{LEMON (200M)} & \textbf{ESM-2 (650M)} & \textbf{ProtTucker (3B)} \\
\midrule
Circular Permutation & AUROC & \textbf{0.779} & 0.476 & 0.720 \\
                     & AUPRC & \textbf{0.285} & 0.081 & 0.238 \\
\midrule
EC Number            & F-max & 0.821 & 0.800 & \textbf{0.878} \\
\midrule
GO Mol.\ Function    & F-max & \textbf{0.639} & 0.603 & 0.635 \\
                     & AUPR  & \textbf{0.625} & 0.584 & 0.616 \\
\midrule
Domain Boundary      & NDO   & 0.758 & \textbf{0.771} & 0.750 \\
                     & NRes  & 0.637 & 0.658 & \textbf{0.674} \\
\bottomrule
\end{tabular}
\end{table}
% \vspace{2pt}
% \raggedright\footnotesize
% Circ.\ Perm.: CIRPIN SCOPe40 (1{,}967 CP / 21{,}160 neg.).
% EC: ProteInfer random split (17{,}398 seqs).
% GO MF: CAFA5 (72{,}390 train / 8{,}002 test).
% Domain: CATH 13{,}198 multi-domain chains, oracle $k$.
% ESM-2 CP AUROC\,$<$\,0.5 because all pairwise similarities exceed 0.98.

We evaluate LEMON's embedding quality across four structurally and functionally distinct tasks: i) Circular Permutation, ii) EC Number Classification, iii) GO Molecular Function, and iv) Domain Boundary Detection; spanning both fold-level and functional retrieval challenges. In all settings, embeddings are \emph{frozen} and no task-specific parameters are learned or tuned, providing a strict measure of zero-shot generalization. Against our primary baselines, ESM-2 and ProtTucker, LEMON achieves competitive or superior performance across tasks, demonstrating that the ZEST contrastive objective yields embeddings with broad structural and functional utility beyond the training objective.

\textbf{Circular Permutation:} Nature has a tendency of reusing successful patterns. A circular permutation (CP) where proteins share highly similar global folds but differ in the ordering of their structural elements completely reorders sequence. Detection therefore requires the embedding to be invariant to linear sequence order—a property sequence statistics alone cannot provide. We evaluate on the CIRPIN SCOPe40 benchmark~\cite{kolodziej2025cirpin}, which comprises 1\,968 verified CP pairs drawn from ASTRAL SCOPe~2.08 at 40\% maximum identity, paired with hard false-positive negatives (similar secondary-structure content but no true CP relationship) and random non-CP pairs. 
% Classification is purely by cosine similarity between frozen embeddings—no threshold tuning. 
ESM-2's near-random prediction is sensitive to sequence order and collapse on CPs as the model was never trained on contrastive representations; LEMON achieves substantially above random despite taking only amino-acid sequences as input. The current frontier is held by structure-based methods: CIRPIN recovers 9/11 pairs on the adjusted benchmark, and TM-align-CP~\cite{zhang2005tm} recovers 7/11, but both require predicted or experimental 3D-coordinates. LEMON sets the state of the art among \emph{sequence-only} methods.

\textbf{EC Number Classification:}
We follow the zero-shot evaluation protocol of ProteInfer~\cite{10.7554/eLife.80942}, using its 12-shard random-split SwissProt test set.
% For each test protein we predict EC labels via leave-one-out cosine $k$-NN ($k{=}15$) within the test set, transferring hierarchically-expanded EC labels from the $k$ nearest neighbors. $F_\mathrm{max}$ reports the best micro-averaged F1 over all confidence thresholds; Precision is recorded at the same optimal threshold.
The supervised ceiling is ProteInfer itself ($F_\mathrm{max}{=}0.977$), trained exclusively on this EC annotation task. Among zero-shot embedding methods, ProtTucker shows dominance followed by LEMON and ESM-2. Indicating its contrastive objective produces a neighborhood structure better aligned with enzymatic function even without any EC supervision.

% LEMON being 15$\times$ smaller yet being competitive to others indicates LEMON's contrastive objective produces a neighborhood structure better aligned with enzymatic function even without any EC supervision.

\textbf{GO Molecular Function:}
We evaluate on the MZSGO benchmark~\cite{10.1093/bioinformatics/btag168}, which uses the CAFA5~\cite{cafa-5-protein-function-prediction} train/test split from SwissProt. Predictions are generated by transferring GO term scores from the $k$ nearest training-set neighbors of each test protein by cosine similarity.
We report $F_\mathrm{max}$ and AUPR over the MF ontology. LEMON and ProtTucker are essentially indistinguishable on this task, which is expected:
GO MF encompasses broad biochemical vocabulary learned from sequence co-evolution at scale, and the LEMON paired with ZEST trained on sequence-level TEDLH hierarchy—is not specifically aligned with GO term boundaries. 

\textbf{Domain Boundary Detection:}
We evaluate on the CATH-17287 benchmark~\cite{lau2023merizo}, comprising 17287 multi-domain PDB~\cite{berman2000protein} chains with CATH boundary annotations, following the Merizo evaluation protocol. Using a sliding window approach, we cluster representations to determine domain counts.
% Embeddings are extracted over dense overlapping windows; a pairwise cosine-similarity matrix is constructed and spectrally clustered into $k$ domains (based on silhouette score) to produce per-residue domain assignments.
NDO (Normalized Domain Overlap) measures how well predicted segments cover each ground-truth domain; NRes measures per-residue assignment accuracy.
LEMON was finetuned on single domain sequence choppings, hence this task is truly novel for our model. LEMON, ESM-2 \& ProtTucker stays at a competitive range. The structure-based frontier is Merizo (NDO~$\approx$0.83), which uses predicted 3D coordinates; the sequence-only frontier prior to this work was Chainsaw~\cite{10.1093/bioinformatics/btae296} (NDO~$\approx$0.72). LEMON exceeds Chainsaw while requiring no structural input.

% NOTE: Zero shot means the model was not explicitly trained on this task, the model just happend to learn 
% \begin{itemize}
%     \item Circular Permutation: define the testing conditions and talk about cirpin
%     \item EC classification: 
%     \item Multi domain detection:
%     \item Something at sequence head level 
%     \item Tables and compare the previously defined models 
%     \item Figure ?
        
% \end{itemize}

\section{Discussion}
\raggedbottom
\textbf{Limitations:} This work has limitations. First, although LEMON never takes structure as an input modality, it receives indirect structural supervision: the CATH hierarchy and TED domain boundaries used during contrastive finetuning are themselves derived from protein structures. In addition, LEMON was finetuned on single-domain sequences, so its behavior on multi-domain proteins and on shorter, motif-level queries remains untested. The zero-shot domain-boundary results in Table~\ref{tab:zeroshot} suggest some transfer, but they do not replace a dedicated multi-domain retrieval evaluation.

Second, ZEST's substitution-invariant vocabulary trades fine-grained detail for fold-level robustness. LEMON's gains are strongest at the fold level, whereas at the superfamily level residue level PLMs such as TM-Vec and larger contrastive models such as ProtTucker remain ahead. We hypothesize that the same trade-off underlies LEMON's weaker residue-level property prediction relative to ESM-2 see appendix~\ref{app:res-head}. While we show that ZEST compression preserves amino acid identities.
 
Third, the bootstrap intervals we report for our tokenizer comparison and ablations capture variability over evaluation queries, not variability across independent training runs. Due to limited compute, the full-scale models were trained with a single seed, and we did not scale beyond 200M parameters. Extending ZEST-style vocabularies to residue-level, multi-domain and multimodal settings, including structure prediction, remains future work.

\textbf{Conclusion:} We introduce ZEST, a novel biologically-grounded protein vocabulary derived from nature's own conservation signals, and demonstrate its effectiveness through LEMON, a compact protein language model trained entirely on a single GPU. Rather than following the prevailing paradigm of scaling parameters and computational resources, we show that embedding evolutionary knowledge directly at the tokenization stage is a novel and efficient alternative.

Proteins carry billions of years of evolution in their sequences, and rather than letting a model slowly rediscover these patterns through massive training, ZEST injects them directly, offering a clear philosophy that domain knowledge belongs at the foundation of a model, not as an afterthought, with broad implications for modern bioinformatics. The advantage of this approach extends well beyond benchmarks: remote homology detection, protein engineering, and some of the deepest open questions in evolutionary biology, including how life reuses successful structural solutions across vastly different species and, more fundamentally, how the universe of protein folds first arose and diversified over evolutionary trajectories.

A key technical ingredient behind ZEST's success is trie-dropout. Our ablation studies show that without dropout, a subset of tokens is rarely or never attended to, leaving their embeddings effectively random. Trie-dropout ensures a more uniform and expressive token utilization across the vocabulary, thus acting as a powerful regularizer and promoting uniform vocabulary coverage.

We designed LEMON, driven by ZEST, for fold-level understanding. LEMON converges faster than its baselines and demonstrates strong zero-shot generalization at the fold level — learning structural relationships directly from sequence space alone, without structural modality. Our results suggest that evolutionary conservation, injected at the tokenization stage, lets a sequence-only model recover fold-level information. 

We release this work as an open proof of concept, and expect that it will be expanded in the future by domain experts in structural and evolutionary biology. We believe that the intersection of biological prior knowledge and representation learning is a largely uncharted territory, and that biologically informed tokenization is one of the most underexplored levers available to the field. Extending ZEST-style vocabularies to residue-level, multi-domain and multi-modal settings, including structure prediction, remains future work.

\textbf{Acknowledgments:} The authors are thankful to  Prof. David Sherrill for the help with computational resources and Prof. Loren Dean Williams for administrative oversight. This work was supported by a cyber-infrastructure grant from the Institute for Data Engineering and Science (IDEaS) at the Georgia Institute of Technology, using resources from the Partnership for an Advanced Computing Environment (PACE, RRID:SCR\_027619). A.S.P was supported by NASA Grant 80NSSC24K0344, C.A.-C by Royal Society Newton International Fellowship NIF\textbackslash{}R1\textbackslash{}231037, and BB was in part supported by fellowship from MSBINF program at Georgia Tech. 

\textbf{Competing interests:} The authors declare no competing interests.

\bibliographystyle{unsrtnat}
\bibliography{references}

% \input{bodies/refs}
%%%%%%%%%%%%%%%%%%%%%%%%%%%%%%%%%%%%%%%%%%%%%%%%%%%%%%%%%%%%
% APPENDIX
%%%%%%%%%%%%%%%%%%%%%%%%%%%%%%%%%%%%%%%%%%%%%%%%%%%%%%%%%%%%
\newpage

\etocsetstyle{section}
  {\par\addvspace{0.5ex}\noindent}
  {\leftskip=0pt \rightskip=2em \parfillskip=-\rightskip}
  {\textbf{\etocnumber\quad\etocname}\ \dotfill\ \etocpage\par}
  {}
\etocsetstyle{subsection}
  {}
  {\leftskip=1.5em \rightskip=2em \parfillskip=-\rightskip}
  {\etocnumber\quad\etocname\ \dotfill\ \etocpage\par}
  {}

\appendix
\begingroup
\makeatletter
\renewcommand{\@setref}[3]{} 
\def\@part[#1]#2{%
  \refstepcounter{part}%
  \addcontentsline{toc}{part}{\thepart\hspace{1em}#1}%
  {\raggedright
   \interlinepenalty \@M
   \normalfont
   \huge \bfseries #2\par}%
  \nobreak
  \vskip 3ex
  \@afterheading}
\makeatother

\part{Appendix}        
\localtableofcontents    
\newpage

\section{Vocabulary Examples}
\subsection{Creation of the ZEST Vocabulary}
\label{app:zest}
% ZEST Algorithm - NeurIPS Style
% Usage: \input{zest_algorithm.tex}
% Requires: algorithm, algpseudocode, amsmath

\begin{algorithm}[h]
\caption{ZEST Vocabulary Construction}
\label{alg:zest}
\begin{algorithmic}[1]
\Require Consensus sequences $\mathcal{S}$ with gap positions marked as \texttt{x}
\Require Min Zone length $\ell_{\min}{=}3$, vocab size $N$
\Ensure Vocabulary $\mathcal{V}$
\Statex \textit{// Phase 1: Zone Extraction}
\For{consensus $S \in \mathcal{S}$}
    \State Extract maximal contiguous substrings where $S[i] \neq \texttt{x}$
    \State Add substrings with length $\geq \ell_{\min}$ to multiset $\mathcal{I}$
\EndFor
\Statex \textit{// Phase 2: Filtering}
\State $\mathcal{I}' \gets \{I \in \mathcal{I} : \text{count}(I) \geq 2 \land |\text{unique}(I)| > 1\}$
\Statex \textit{// Phase 3: Clustering with 70\% identity}
\State $\mathcal{C} \gets \textsc{LinClust}(\mathcal{I}', \theta{=}0.7)$ 
\Statex \hspace{1em} where $\mathcal{C} = \{(r_1, M_1), \ldots, (r_k, M_k)\}$,
\Statex \hspace{1em} $r_i$ = cluster representative, $M_i$ = member Zones
\Statex \textit{// Phase 4: Scoring \& Selection}
\For{cluster $(r, M_r) \in \mathcal{C}$}
    \State $\text{score}(r) \gets \left(\sum_{m \in M_r} \text{count}(m)\right) \cdot |r|^{1.5}$
\EndFor
\State $\mathcal{V} \gets \text{Top-}N(\mathcal{C}, \text{score})$
\State \Return $\mathcal{V}$
\end{algorithmic}
\end{algorithm}

We capture the census sequence from MSA profiles from TEDLH data~\cite{carreno2026tedlh},
which is publicly available, and extract zones using Algorithm~\ref{alg:zest}.
Each member of a cluster is assigned the same token id to address sequence mutability.
Smaller zones are highly frequent and more easily clustered; therefore, to prevent our vocabulary from becoming excessively skewed toward small motifs and to ensure longer motifs are not penalized by their naturally lower raw frequencies we bias our score slightly to reward length.
\begin{equation*}
\text{score} = \text{freq} \cdot \text{length}^{1.5}
\end{equation*}
Running Algorithm~\ref{alg:zest} over the TEDLH census sequences yields a closed vocabulary of \textbf{31,975 tokens}.
Of these, 27,524 (86.1\,\%) are \emph{singletons}---zones for which no biochemically interchangeable variant exists in the corpus and which therefore receive a unique token identity.
The remaining 4,451 tokens (13.9\,\%) are genuine \emph{clusters}: each maps two or more mutually substitutable sequence variants to a single token id, collectively encoding 43,388 distinct sequence fragments within the 32K vocabulary budget.

The median cluster size is 3--4 members, consistent with the degeneracy expected from conservative single-position substitutions (e.g.\ D$\leftrightarrow$N$\leftrightarrow$E, I$\leftrightarrow$V$\leftrightarrow$L).
Token lengths span 3 to 82 residues with a median of 4---providing a context compression of ${\sim}4{\times}$ relative to single-residue tokenization while preserving multi-residue zones as atomic sequence units.

% \newpage
% \subsection{Example ZEST Vocabulary Clusters}
% \input{bodies/tables/clusters}
\newpage

\subsection{Test Time Augmentation:}
\begin{algorithm}[H]
\caption{Trie-Dropout Tokenization}\label{alg:trie-dropout}
\begin{algorithmic}[1]
\Require Symbol sequence $\mathbf{x} = (x_1, \dots, x_N)$, trie $\mathcal{T}$, dropout rate $d \in [0, 1]$
\Ensure Token ID sequence $\mathbf{t} = (t_1, \dots, t_M)$
\State $\mathbf{t} \gets [\,]$;\; $i \gets 1$
\While{$i \leq N$}
    \State $\mathcal{M} \gets \Call{TrieMatch}{\mathcal{T},\, \mathbf{x},\, i}$
    \If{$\mathcal{M} = \emptyset$}
        \State append \textsc{[mask]} to $\mathbf{t}$;\; $i \gets i + 1$;\; \textbf{continue}
    \EndIf
    \State $(\ell^*,\, t^*) \gets \mathcal{M}[0]$
    \If{$d > 0$ \textbf{and} $\mathrm{Uniform}(0,1) < d$ \textbf{and} $|\mathcal{M}| > 1$}
        \State $j \sim \mathrm{Uniform}\{1, \dots, |\mathcal{M}|-1\}$
        \State $(\ell^*,\, t^*) \gets \mathcal{M}[j]$
    \EndIf
    \State append $t^*$ to $\mathbf{t}$;\; $i \gets i + \ell^*$
\EndWhile
\State \Return $\mathbf{t}$
\end{algorithmic}
\end{algorithm}

% ============================================================
% Algorithm 2 — Test-Time Augmentation with Trie-Dropout
% ============================================================
\begin{algorithm}[H]
\caption{Test-Time Augmentation with Trie-Dropout}\label{alg:tta}
\begin{algorithmic}[1]
\Require Sequence $\mathbf{x}$, encoder $f_\theta$, trie $\mathcal{T}$, dropout rate $d$, passes $K$
\Ensure Embedding $\bar{\mathbf{e}} \in \mathbb{R}^D$
\For{$k = 1, \dots, K$}
    \State $\mathbf{t}^{(k)} \gets \Call{TrieDropoutTokenize}{\mathbf{x},\, \mathcal{T},\, d}$
        \Comment{Alg.~\ref{alg:trie-dropout}}
    \State $\mathbf{e}^{(k)} \gets f_\theta\!\bigl(\mathbf{t}^{(k)}\bigr)$
\EndFor
\State $\bar{\mathbf{e}} \gets \frac{1}{K} \sum_{k=1}^{K} \mathbf{e}^{(k)}$
\State \Return $\bar{\mathbf{e}}$
\end{algorithmic}
\end{algorithm}
\newpage

\section{Training}
\label{app:training_details}
\subsection{Training Objective}
\textbf{Pretraining with Masked Language Model:}
We pre-train the encoder and sequence head with a dual-objective masked language model. Adapting the BERT masking convention~\cite{devlin2019bert} to protein domain, we mask 30\% of input tokens that are selected for prediction: 80\% are replaced with \texttt{[MASK]}, 10\% with a random vocabulary token, and 10\% are left unchanged. The aggressive 30\% masking rate (vs.\ BERT's standard 15\%) is motivated by the low information density of protein sequences.

\textbf{Dual MLM objective.}
The model jointly optimizes two heads: (i)~a token-level MLM head that predicts the masked ZEST token via tied input embeddings, and (ii) a residue-level MLM head (\texttt{ExpansionHead}) that expands each token embedding to per-residue amino acid predictions, predicting the identity of each constituent amino acid within a masked token. The token-level task operates over a 32K vocabulary with $\sim$4--6$\times$ compression, resulting in high per-token perplexity (tokens encode multi-residue zones, making prediction inherently harder than single-residue prediction). The residue-level task operates over the 20 standard amino acids. The two losses are combined via learned uncertainty weighting\cite{kendall2018multi}: each task $i$ has a learnable log-variance $s_i$, and the combined loss is $\mathcal{L} = \sum_i \left( w_i \cdot e^{-s_i} \cdot \mathcal{L}_i + \tfrac{1}{2} s_i \right)$, with priority weights $w_\text{tok}{=}0.7$ and $w_\text{res}{=}0.3$. This allows the model to automatically down weight the noisier objective over the course of training.

\textbf{Trie dropout.}
At each epoch, sequences are re-tokenized with 15\% trie-level dropout: during greedy trie matching, each node extension is randomly dropped with probability $p{=}0.15$, producing a different segmentation of the same sequence at every epoch.
This acts as a powerful data augmentation that prevents the model from memorizing fixed token boundaries and forces robustness to segmentation variability.

\textbf{Pretraining results.}
The dual-MLM task is intentionally harder than single-residue prediction: the 32K ZEST vocabulary means each masked token encodes a variable-length characteristic signature (mean ${\sim}4$ residues), so the token-level prediction space is orders of magnitude larger.
Our model reaches token perplexity $1{,}098$ and $9.6$ residue perplexity. The high token-level perplexity reflects the combinatorial difficulty of the task---predicting which of 32K multi-residue motifs was masked---rather than poor model quality; the strong residue-level accuracy confirms that the encoder learns meaningful amino acid context.
% ------------------------------------------------------------------
\subsection{Finetuning: Hierarchical Contrastive Learning}
In the second phase, we fine-tune on CATH-derived protein pairs with hierarchical contrastive learning.
An attention-based aggregator pools variable-length token embeddings into a fixed-dimensional sequence representation, which is mapped through a multi-layer bottleneck projector and $\ell_2$-normalized.

\textbf{Hierarchical InfoNCE.}
The primary objective is a weighted sum of three per-level InfoNCE~\cite{oord2018representation} losses operating at the Architecture ($\tau_A{=}0.10$, $w_A{=}0.15$), Topology ($\tau_T{=}0.07$, $w_T{=}0.35$), and Superfamily ($\tau_H{=}0.05$, $w_H{=}0.50$) levels.
Hard negatives---pairs sharing a coarser CATH label but differing at the target level---receive $2{\times}$ up-weighted logits in the InfoNCE denominator, and a FIFO memory bank detached embeddings provides additional negatives beyond the current batch.

\subsection{Sequence head as regularizer}
\label{app:dual-head}
\begin{figure}[htbp]
    \centering
    \includegraphics[width=\textwidth]{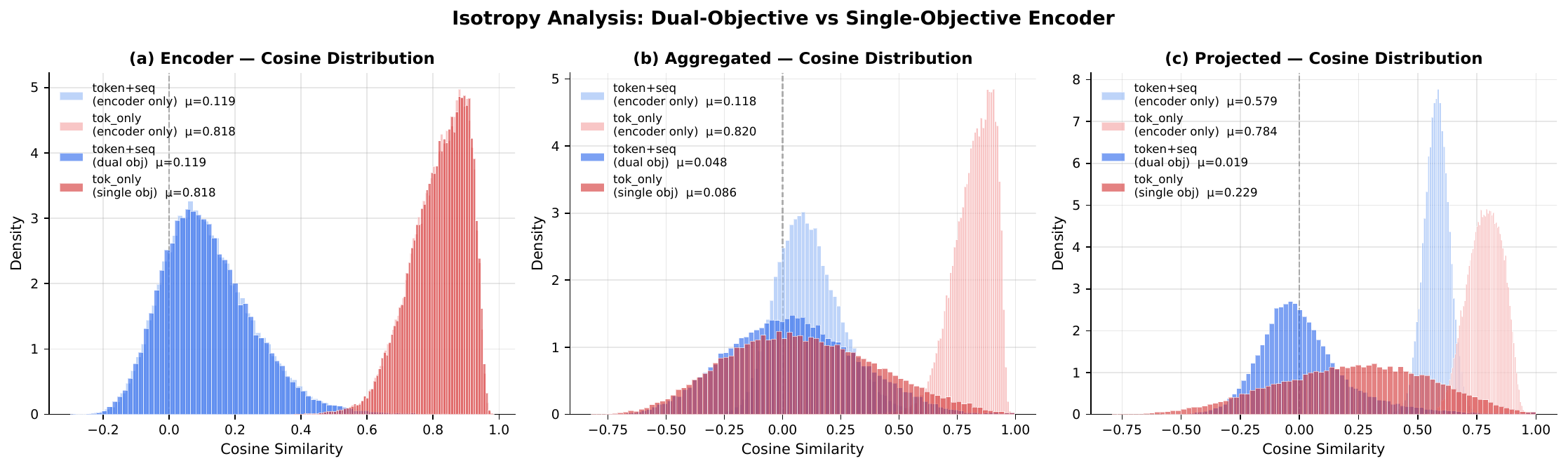}
    \caption{Pairwise cosine-similarity distributions at three representation levels---encoder output (before pooling), aggregated sequence embedding, and projected contrastive space---for LEMON (blue) and Tok-only (red) in pretrained and finetuned states. A distribution centred near zero with low variance indicates isotropic, evenly-spread representations; a peak near $+1$ indicates cone collapse. LEMON's dual-objective backbone emerges from pre-training already well-spread at the encoder level ($\bar{\cos}{=}0.12$), and contrastive fine-tuning compounds this advantage, pushing the projected distribution to near-zero mean ($\bar{\cos}{=}0.02$). The Tok-only backbone, despite identical fine-tuning, cannot overcome its anisotropic initialization ($\bar{\cos}{=}0.82$) and retains a substantially higher projected mean ($\bar{\cos}{=}0.23$), consistent with the observed precision ceiling.}
    \label{fig:isotropy}
\end{figure}
\begin{table*}[t]
\centering
\small
\setlength{\tabcolsep}{4.5pt} % Tighten column padding to prevent horizontal overfull
\caption{Intrinsic representation geometry (isotropy) and downstream structural classification performance. $\bar{\cos}$: mean pairwise cosine similarity (lower = more isotropic). $d_{\text{eff}}$: effective dimensionality. $k_{90}$: principal components for 90\% variance. LEMON: dual-head; Tok-only: single-head.}
\label{tab:isotropy_downstream}
\begin{tabular}{ll ccc cccc}
\toprule
& & \multicolumn{3}{c}{\textbf{Isotropy Metrics}} & \multicolumn{2}{c}{\textbf{Fold / Arch}} & \multicolumn{2}{c}{\textbf{Superfam / Topo}} \\
\cmidrule(lr){3-5} \cmidrule(lr){6-7} \cmidrule(lr){8-9}
\textbf{Model} & \textbf{Level} & $\bar{\cos} \downarrow$ & $d_{\text{eff}} \uparrow$ & $k_{90} \uparrow$ & AUROC $\uparrow$ & mAP $\uparrow$ & AUROC $\uparrow$ & mAP $\uparrow$ \\
\midrule
\multicolumn{9}{l}{\textit{\textbf{Pretrained}}} \\
LEMON    & Encoder & $+0.119$ & $43.0$ & $183$ & $0.636$ & $0.051$ & $0.775$ & $0.104$ \\
Tok-only & Encoder & $+0.818$ & $19.5$ & $119$ & -- & -- & -- & -- \\
\midrule
\multicolumn{9}{l}{\textit{\textbf{Finetuned}}} \\
\multirow{2}{*}{LEMON}    & Encoder          & $+0.119$ & $43.0$ & $183$ & $0.758$ & $0.173$ & $0.882$ & $0.324$ \\
                          & Contrastive-head & $+0.019$ & $25.4$ & $29$  & $\mathbf{0.875}$ & $\mathbf{0.316}$ & $\mathbf{0.927}$ & $\mathbf{0.405}$ \\
\addlinespace
\multirow{2}{*}{Tok-only} & Encoder          & $+0.818$ & $19.5$ & $119$ & $0.678$ & $0.117$ & $0.802$ & $0.183$ \\
                          & Contrastive-head & $+0.229$ & $8.6$  & $9$   & $0.825$ & $0.192$ & $0.921$ & $0.329$ \\
\bottomrule
\end{tabular}
\end{table*}
\paragraph{Isotropy analysis.}
Early experiments trained a single-objective encoder (token-level masked language
modeling only) and fine-tuned it with a contrastive retrieval head.
Although this model achieved competitive fold-level AUROC, precision (mAP / AUPRC)
remained consistently poor and could not be recovered through further training,
learning-rate tuning, or temperature annealing.
We hypothesized that the root cause was representation collapse in the encoder itself:
a token-only pretraining signal provides no pressure for the encoder to distribute
information across the full embedding space.
To verify this, we measured representation isotropy at three levels---encoder
output (before pooling), aggregated sequence embedding, and projected
contrastive embedding---for both the pretrained backbones and their
fine-tuned counterparts.
Below the figure shows the pairwise cosine-similarity distributions
underlying these measurements; a distribution centred near zero with low
variance indicates isotropic, evenly-spread representations, whereas a peak
near $+1$ indicates anisotropy (cone collapse).
The token-only pretrained backbone exhibits severe anisotropy at the encoder
level ($\bar{\cos} = +0.818$, Table~\ref{tab:isotropy}), meaning nearly all
representations point in the same direction before any fine-tuning has occurred.
Even after full end-to-end contrastive fine-tuning, this pathology persists in
the projected space ($\bar{\cos} = +0.229$), consistent with the observed
precision ceiling.
By contrast, the dual-objective backbone, trained jointly on token MLM and
residue-level reconstruction, emerges from pre-training with a substantially
more isotropic encoder ($\bar{\cos} = +0.119$); the residue-level signal
acts as a regularizer that prevents the encoder from collapsing to a
low-dimensional subspace.
After fine-tuning, this advantage compounds: the aggregated embedding reaches
$\bar{\cos} = +0.048$ and the projected embedding reaches $\bar{\cos} = +0.019$
(Panel~(c) of Figure~\ref{fig:isotropy}), near-uniform coverage of the
hypersphere, directly enabling the precision gains reported in
Table~\ref{tab:isotropy}.
This analysis motivated the dual-head architecture: the residue MLM head is not
merely an auxiliary loss but is structurally necessary to produce encoder
representations from which high-precision retrieval is recoverable.

% \input{bodies/6.Appendisx/training}
% \subsection{Hardware}
% \input{bodies/tables/trie_algo}

\section{Task Evaluations}
\subsection{Remote Homology Inference}

\paragraph{Embedding extraction.}
Each protein sequence is tokenized with ZEST (greedy max-match, trie-dropout disabled at test time), passed through the shared 24-layer encoder, pooled by the learned-query aggregator, and projected to a 384-dimensional $\ell_2$-normalized vector.
For TTA experiments ($K{>}1$), each sequence is tokenized $K$ independent times at the specified dropout rate $d$; the resulting $K$ embeddings are averaged and re-normalized before retrieval.

\paragraph{Pair labeling.}
Ground-truth labels follow the SCOP and CATH classification hierarchies.
At target level $\ell$, a candidate $c$ relative to query $q$ is classified as:
\begin{itemize}[noitemsep]
    \item \textbf{Positive}: $q$ and $c$ agree at level $\ell$ but differ at the immediately lower level.
    \item \textbf{Negative}: $q$ and $c$ disagree at level $\ell$.
    \item \textbf{Ignored}: $q$ and $c$ agree at both levels.
\end{itemize}
Ignored pairs are excluded from all metric computations.
This definition ensures that positive pairs span exactly one level of the hierarchy---for example, at the SCOP Fold level, two sequences must share a fold but belong to different superfamilies.

\subsection{Residue Head Evaluations}

\paragraph{Introduction:} While the main purpose is to help regularize the training, in this section we try to evaluate the sequence heads performance against ESM2.

\paragraph{Masked reconstruction under levels of masking.} We compare LEMON with ESM-2 650M on 4,000 held-out UniRef90 windows (mean length 83.8 residues), scoring both models on the same hidden residue positions. We report amino-acid perplexity over scored residues, averaged over three mask seeds, with 95\% bootstrap intervals over sequences. The models are evaluated on two masking schemes, i) \emph{token-span} masking, contiguous spans of 3--8 ZEST tokens are hidden and every residue they cover is scored and ii) \emph{residue-target} masking, residue sites are sampled at the given rate, every ZEST token overlapping a sampled site is masked, and only the sampled sites are scored. Because a ZEST token spans several residues, the fraction of residues actually hidden exceeds the nominal rate, most strongly under residue-target masking, so the two schemes are not comparable at the same column. At 0\% nothing is masked, so the column measures reconstruction from the tokens alone, and ESM-2 is better there and under light masking. As masking increases, ESM-2 degrades sharply under token-span masking while LEMON stays nearly flat, with LEMON overtaking ESM-2 at moderate rates. Under residue-target masking the crossover comes earlier, and ESM-2 becomes non-monotonic because the scored sites are a shrinking fraction of an almost fully hidden sequence. LEMON was pretrained with a 30\% masking rate, so this range includes its training condition. Its perplexity stays near a composition-level floor (a uniform guess over 20 amino acids gives 20), so the result shows robustness to heavy context loss rather than retained residue-level information. 

\begin{table}[!h]
\centering
\caption{AA-reconstruction perplexity ($\downarrow$) on UniRef90 ($N{=}4000$) under increasing masking.}
\label{tab:masking_ppl}
\small
\renewcommand{\arraystretch}{1.5}
\setlength{\tabcolsep}{10pt}
\begin{tabular}{@{}llccccccc@{}}
\toprule
 & & \multicolumn{7}{c}{\textbf{Masking rate (\%)}} \\
\cmidrule(l){3-9}
\textbf{Masking} & \textbf{Model} & \textbf{0} & \textbf{10} & \textbf{20} & \textbf{30} & \textbf{40} & \textbf{50} & \textbf{60} \\
\midrule
\multirow{2}{*}{Token-span}
& LEMON & 1.94 & 16.12 & 16.26 & 16.47 & \textbf{16.61} & \textbf{16.75} & \textbf{16.88} \\
& ESM-2 & \textbf{1.45} & \textbf{12.53} & \textbf{13.16} & \textbf{15.71} & 23.33 & 50.54 & 137.02 \\
\addlinespace[4pt]
\midrule
\addlinespace[4pt]
\multirow{2}{*}{Residue-target}
& LEMON & 1.94 & 15.29 & \textbf{15.99} & \textbf{16.37} & \textbf{16.67} & \textbf{17.02} & \textbf{17.49} \\
& ESM-2 & \textbf{1.45} & \textbf{10.87} & 21.17 & 88.64 & 92.70 & 71.30 & 59.20 \\
\bottomrule
\end{tabular}
\end{table}

\label{app:res-head}

\paragraph{PPI Interface Site Prediction.}
\begin{figure}[h]
  \centering
  \includegraphics[width=\linewidth]{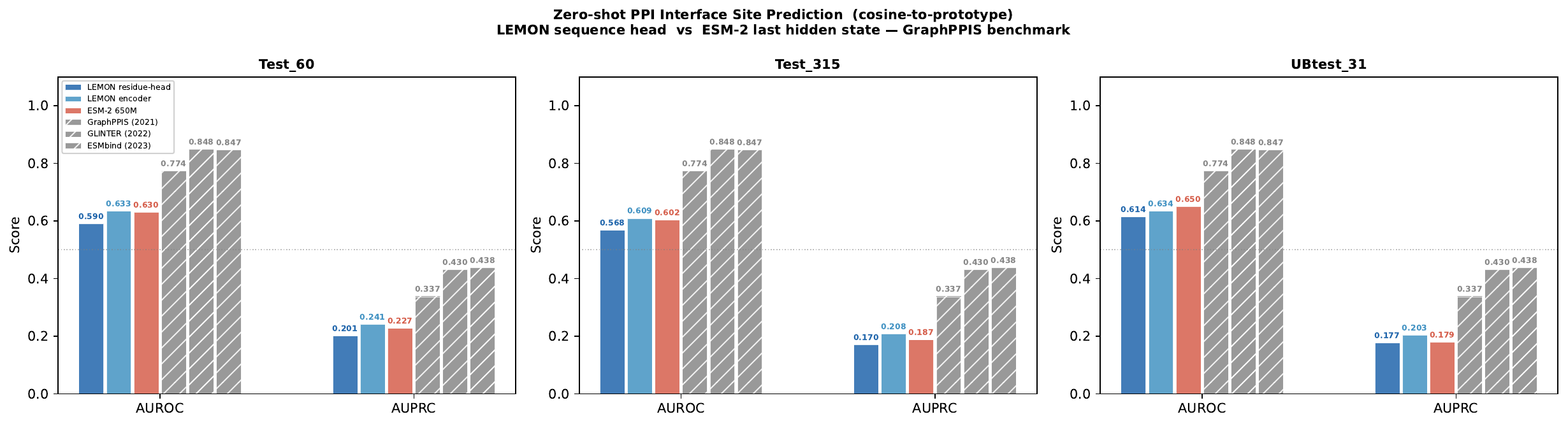}
  \caption{PPI interface site prediction results on GraphPPIS.}
  \label{fig:ppi}
\end{figure}
We evaluate zero-shot on the GraphPPIS benchmark~\cite{gao2023hierarchical}
across three test splits: \textsc{Test\_60} (60 chains), \textsc{Test\_315}
(315 chains), and \textsc{UBtest\_31} (31 unbound structures).
A prototype interface embedding is constructed by mean-pooling per-residue
representations from the residue expansion head over all labeled interface
residues in \textsc{Train\_335}; at inference each residue is scored by its
cosine similarity to this prototype.
The LEMON encoder variant achieves AUROC $0.633$ / AUPRC $0.241$ on
\textsc{Test\_60} and AUROC $0.609$ / AUPRC $0.208$ on \textsc{Test\_315},
marginally ahead of ESM-2 650M ($0.630$ / $0.227$ and $0.602$ / $0.187$,
respectively).
The residue-head variant scores lower at $0.590$ / $0.201$ and $0.568$ /
$0.170$.
Supervised task-specific models remain substantially ahead
(GraphPPIS~\cite{gao2023hierarchical} AUROC $0.774$;
 as expected under a zero-shot protocol.

\paragraph{Thermo-stability Prediction.}
\begin{figure}[!h]
  \centering
  \includegraphics[width=\linewidth]{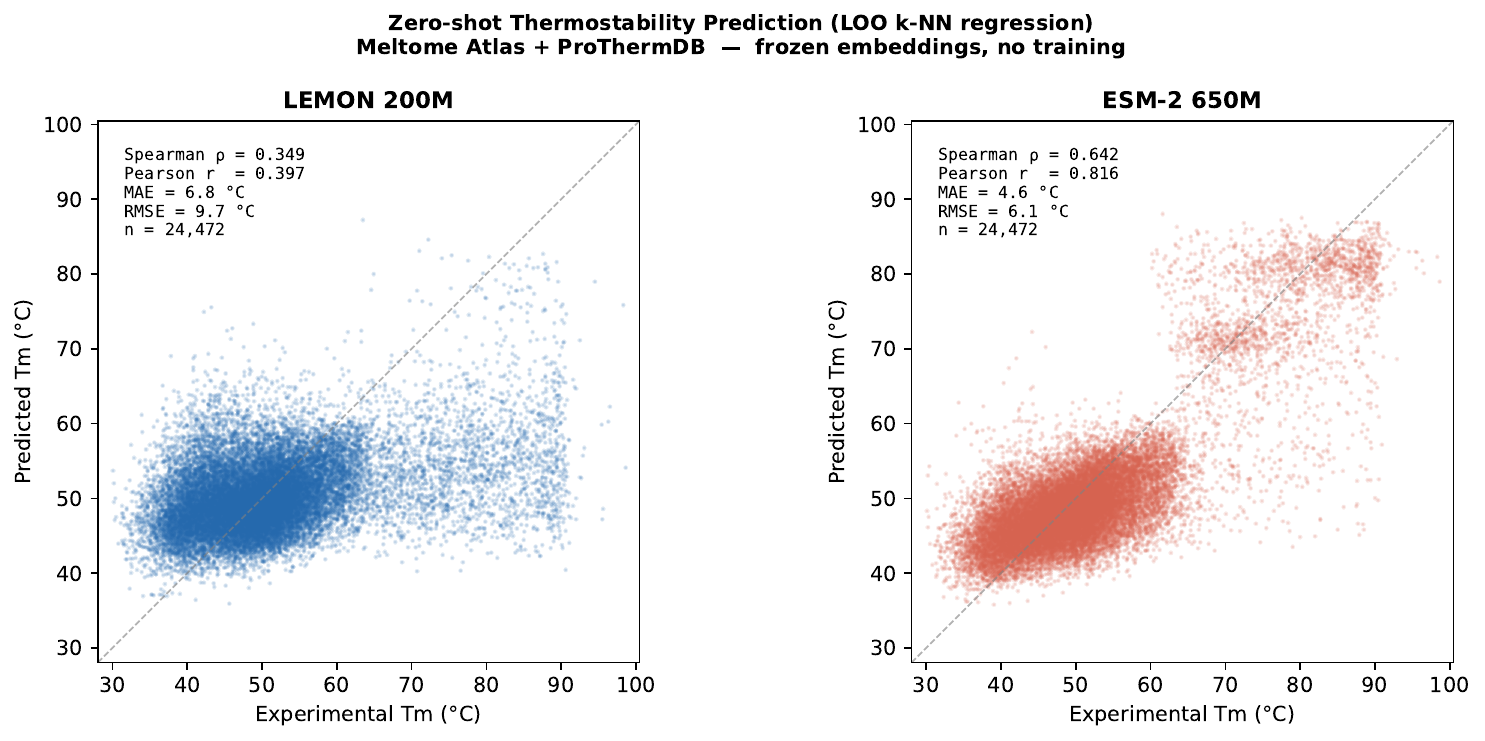}
  \caption{Zero-shot thermostability prediction on the Meltome Atlas.}
  \label{fig:thermo}
\end{figure}

We evaluate zero-shot on the Meltome Atlas~\cite{jarzab2020meltome} (${\sim}24{,}472$ proteins with experimental melting temperatures $T_m$,
filtered at 70\% sequence identity). Sequences are encoded by mean-pooling encoder token representations, and $T_m$ is predicted by leave-one-out $k$-NN regression ($k{=}10$, cosine distance).
LEMON achieves Spearman $\rho = 0.349$, Pearson $r = 0.397$, MAE $= 6.85\,^{\circ}$C, RMSE $= 9.73\,^{\circ}$C, versus ESM-2 650M at $\rho = 0.642$, $r = 0.816$,
MAE $= 4.59\,^{\circ}$C, RMSE $= 6.13\,^{\circ}$C.
The gap reflects that LEMON is trained with a contrastive fold-similarity
objective rather than sequence reconstruction, making its encoder embeddings
less sensitive to the fine-grained physicochemical signals that drive thermal
stability.

\subsection{Reproduction of Results}
\label{app:reproduction}

All code, model weights, the ZEST tokenizer, and evaluation scripts are publicly available at \href{https://github.com/LDWLab/ZEST}{github.com/LDWLab/ZEST} under the MIT license, with instructions to reproduce every table and figure. Model checkpoints are hosted on Hugging Face.

\paragraph{Computational resources.}
\label{app:hardware}
All training was conducted on a single NVIDIA H100 80\,GB HBM3 GPU
(driver 570.124.06 and torch 2.13.0+cu130), with an Intel Xeon Platinum~8480+ CPU and 221\,GiB system RAM.
Pre-training consumed ${\approx}122$ GPU-hours over 9 epochs
(mean 13.5\,h/epoch; the 3.7\,h epoch-7 scheduler-restart run is excluded from the mean);
the frozen contrastive phase added 1.7\,GPU-hours;
end-to-end fine-tuning required ${\approx}45$ GPU-hours over 6 epochs
(mean 9.0\,h/epoch).
Total compute for the reported model: ${\approx}169$ GPU-hours on a single H100.
The corrected evaluations and benchmarks were run on NVIDIA L40S gpu with torch version 2.13.0+cu130.

% \paragraph{Computational cost.}
% Embedding all 7\,117 SCOPe sequences takes approximately 2~minutes on a single
% H100 or A100 GPU (\texttt{batch\_size=128}); the SCOP~1.75 corpus
% (31\,073 sequences) requires approximately 8~minutes.
% Pairwise cosine-similarity computation is performed in NumPy and completes in
% under 30~seconds for all three datasets on CPU.
% No GPU is strictly required: the \texttt{-{}-device cpu} flag enables full
% CPU-only evaluation at the cost of roughly $10{\times}$ longer embedding
% extraction.

\end{document}